\documentclass[10pt,twocolumn,letterpaper]{article}

\usepackage{cvpr}
\usepackage{times}
\usepackage{epsfig}
\usepackage{graphicx}
\usepackage{amsmath}
\usepackage{amssymb}

\usepackage{cite}
\usepackage{wrapfig}
\usepackage{multirow}
\usepackage{caption}
\usepackage{subcaption}
\usepackage{verbatim}
\usepackage{booktabs}
\usepackage{comment}
\usepackage{mathtools}
\DeclarePairedDelimiter{\floor}{\lfloor}{\rfloor}
\DeclarePairedDelimiter{\ceil}{\lceil}{\rceil}
\DeclareMathOperator*{\argmin}{arg\,min}

\usepackage[pagebackref=true,breaklinks=true,letterpaper=true,colorlinks,bookmarks=false]{hyperref}

\def\cvprPaperID{203} 

\def\eqnvspace{{\vspace{-2mm}}}
\def\figvspace{{\vspace{-4mm}}}

\def\afifthcolumn{0.075\textwidth}
\def\FittingFigWid{0.11\textwidth}

\def\FittingFigShapeWid{0.11\textwidth}
\def\AlignFigWid{0.11\textwidth}
\def\AblTexFigWid{0.1\textwidth}
\def\VaryingShapeFigWid{0.12\textwidth}

\newcommand{\norm}[1]{\left\lVert#1\right\rVert}

\newcommand{\Paragraph}[1]{\vspace{-0mm} \noindent \textbf{#1.} \hspace{0mm}}
\newcommand{\Section}[1]{\vspace{-1mm} \section{#1} \vspace{-0mm}}
\newcommand{\SubSection}[1]{\vspace{-1mm} \subsection{#1} \vspace{-1mm}}
\newcommand{\SubSubSection}[1]{\vspace{-3mm} \subsubsection{#1} \vspace{-1mm}}

\cvprfinalcopy 

\ifcvprfinal\fi
\begin{document}

\title{Nonlinear 3D Face Morphable Model}

\author{Luan Tran, Xiaoming Liu \\
Department of Computer Science and Engineering \\
Michigan State University, East Lansing MI 48824\\
{\tt \{tranluan, liuxm\}@msu.edu}
}

\maketitle

\begin{abstract}
As a classic statistical model of 3D facial shape and texture, 3D Morphable Model (3DMM) is widely used in facial analysis, e.g., model fitting, image synthesis.
Conventional 3DMM is learned from a set of well-controlled 2D face images with associated 3D face scans, and represented by two sets of PCA basis functions.
Due to the type and amount of training data, as well as the linear bases, the representation power of 3DMM can be limited.
To address these problems, this paper proposes an innovative framework to learn a nonlinear 3DMM model from a large set of unconstrained face images, without collecting 3D face scans. 
Specifically, given a face image as input, a network encoder estimates the projection, shape and texture parameters.
Two decoders serve as the nonlinear 3DMM to map from the shape and texture parameters to the 3D shape and texture, respectively.
With the projection parameter, 3D shape, and texture, a novel analytically-differentiable rendering layer is designed to reconstruct the original input face.
The entire network is end-to-end trainable with only weak supervision. 
We demonstrate the superior representation power of our nonlinear 3DMM over its linear counterpart, and its contribution to face alignment and 3D reconstruction.
\footnote{Project page:~\href{http://cvlab.cse.msu.edu/project-nonlinear-3dmm.html}{http://cvlab.cse.msu.edu/project-nonlinear-3dmm.html}}
\end{abstract}


\Section{Introduction}
\label{sec:intro}
3D Morphable Model (3DMM) is a statistical model of 3D facial shape and texture in a space where there are explicit correspondences~\cite{blanz1999morphable}. 
The morphable model framework provides two key benefits: first, a point-to-point correspondence between the reconstruction and all other models, enabling “morphing”,
and second, modeling underlying transformations between types of faces (male to female, neutral to smile, etc.). 
3DMM has been widely applied in numerous areas, such as computer vision~\cite{blanz1999morphable, yin2017towards}, graphics~\cite{aldrian2013inverse}, human behavioral analysis~\cite{amberg2008expression} and craniofacial surgery~\cite{staal2015describing}.

\begin{figure}[t!]
\centering
\includegraphics[trim=0 30 12 0, clip, width=\linewidth]{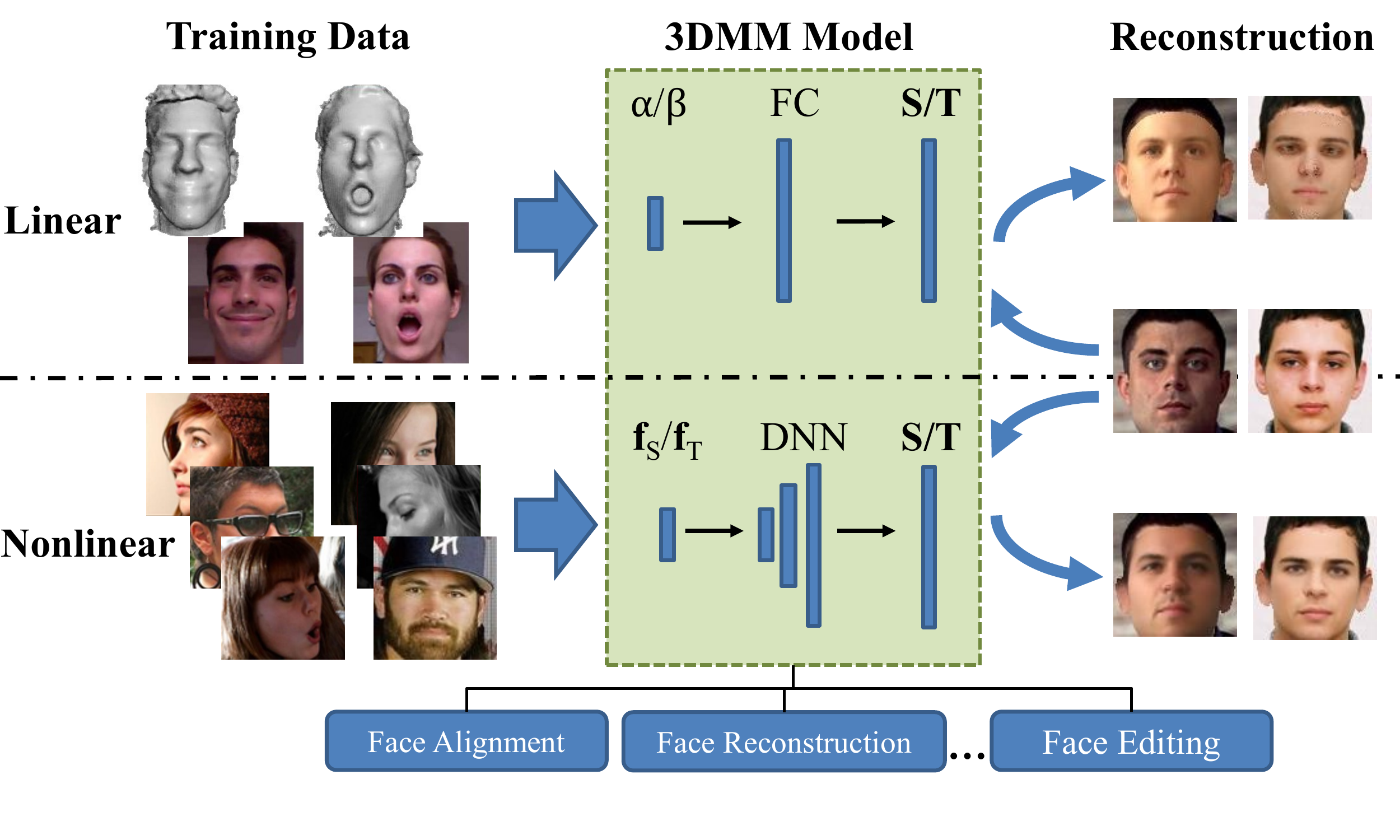}
\vspace{-4mm}
\caption{\small Conventional 3DMM employs linear bases models for shape/texture, which are trained with 3D face scans and associated controlled 2D images. We propose a nonlinear 3DMM to model shape/texture via deep neural networks~(DNNs). It can be trained from in-the-wild face images without 3D scans, and also better reconstructs the original images due to the inherent nonlinearity.}
\label{fig:concept}
\figvspace 
\end{figure}

3DMM is learnt through {\it supervision} by performing dimension reduction, normally Principal Component
Analysis (PCA), on a training set of face images/scans. 
To model highly variable 3D face shapes, a large amount of high-quality 3D face scans is required. 
However, this requirement is expensive to fulfill. 
The first 3DMM~\cite{blanz1999morphable} was built from scans of $200$ subjects with a similar ethnicity/age group. 
They were also captured in well-controlled conditions, with only neutral expressions. 
Hence, it is fragile to large variances in the face identity. 
The widely used Basel Face Model~(BFM)~\cite{paysan20093d} is also built with only $200$ subjects in neutral expressions.  
Lack of expression can be compensated using expression bases from FaceWarehouse~\cite{cao2014facewarehouse} or BD-3FE~\cite{yin20063d}. 
After more than a decade, almost all models use less than $300$ training scans. 
Such a small training set is far from adequate to describe the full variability of human faces~\cite{booth20163d}. 
Only recently, Booth et al.~\cite{booth20163d} spent a significant effort to build 3DMM from scans of ${\sim}10,000$ subjects. 

Second, the texture model of 3DMM is normally built with a small number of 2D face images {\it co-captured} with 3D scans, under well-controlled conditions.
Therefore, such a model is only learnt to represent the facial texture in similar conditions, rather than in-the-wild environments.
This substantially limits the application scenarios of 3DMM.

Finally, the representation power of 3DMM is limited by not only the size of training set but also its formulation. 
The facial variations are nonlinear in nature. 
E.g., the variations in different facial expressions or poses are nonlinear, which violates the linear assumption of PCA-based models. 
Thus, a PCA model is unable to interpret facial variations well.
Given the barrier of 3DMM in its data, supervision and linear bases, this paper aims to revolutionize the paradigm of learning 3DMM by answering a fundamental question:

\begin{quote}
{\it Whether and how can we learn a nonlinear 3D Morphable Model of face shape and texture from a set of unconstrained 2D face images, without collecting 3D face scans?}
\end{quote}

If the answer were yes, this would be in sharp contrast to the conventional 3DMM approach, and remedy all aforementioned limitations.
Fortunately, we have developed approaches that offer positive answers to this question.
Therefore, the core of this paper is regarding how to learn this new 3DMM, what is the representation power of the model, and what is the benefit of the model to facial analysis.

As shown in Fig.~\ref{fig:concept}, starting with an observation that the linear 3DMM formulation is equivalent to a single layer network, using a deep network architecture naturally increases the model capacity. 
Hence, we utilize two network decoders, instead of two PCA spaces, as the shape and texture model components, respectively.
With careful consideration of each component, we design different networks  for shape and texture: the multi-layer perceptron~(MLP) for shape and convolutional neural network~(CNN) for texture.
Each decoder will take a shape or texture representation as input and output the dense 3D face or a face texture.
These two decoders are essentially the nonlinear 3DMM.

Further, we learn the fitting algorithm to our nonlinear 3DMM, which is formulated as a CNN encoder.
The encoder takes a 2D face image as input and generates the shape and texture parameters, from which two decoders estimate the 3D face and texture.
The 3D face and texture would {\it perfectly} reconstruct the input face, if the fitting algorithm and 3DMM are well learnt.
Therefore, we design a differentiable rendering layer to generate a reconstructed face by fusing the 3D face, texture, and the camera projection parameters estimated by the encoder. 
Finally, the end-to-end  learning scheme is constructed where the encoder and two decoders are learnt jointly to minimize the difference between the reconstructed face and the input face.
Jointly learning the 3DMM and the model fitting encoder allows us to leverage the large collection of {\it unconstrained} 2D images without relying on 3D scans.
We show significantly improved shape and texture representation power over the linear 3DMM. 
Consequently, this also benefits other tasks such as 2D face alignment and 3D reconstruction.

In this paper, we make the following contributions:

1) We learn a \textit{nonlinear} 3DMM model that has greater representation power than its traditional linear counterpart.

2) We jointly learn the model and the model fitting algorithm via \textit{weak supervision}, by leveraging a large collection of 2D images without 3D scans. 
The novel rendering layer enables the end-to-end training.

3) The new 3DMM further improves performance in related tasks: face alignment and face reconstruction.

\Section{Prior Work}
\label{sec:prior}

\Paragraph{Linear 3DMM}
Since the original work by Blanz and Vetter~\cite{blanz1999morphable}, there has been a large amount of effort trying to improve 3DMM modeling mechanism.
Paysan et al.~\cite{paysan20093d} use a Nonrigid Iterative Closest Point~\cite{amberg2007optimal} to directly align 3D scans as an alternative to the UV space alignment method in ~\cite{blanz1999morphable}.
Vlasic et al.~\cite{vlasic2005face} use a multilinear model to model the combined effect of identity and expression variation on the facial shape.
Later, Bolkart and Wuhrer~\cite{bolkart2015groupwise} show how such a multilinear model can be estimated directly from the 3D scans using a joint optimization over the model parameters and groupwise registration of 3D scans.

\Paragraph{Improving Linear 3DMM}
With PCA bases, the statistical distribution underlying 3DMM is Gaussian. Koppen et al.~\cite{koppen2017gaussian} argue that single-mode Gaussian can't represent real-world distribution. They introduce the Gaussian Mixture 3DMM that models the global population as a mixture of Gaussian subpopulations, each with its own mean, but shared covariance.
Booth el al.~\cite{booth20173d} aim to improve texture of 3DMM to go beyond controlled settings by learning “in-the-wild” feature-based texture model.
However, both works are still based on statistical PCA bases.
Duong et al.~\cite{nhan2015beyond} address the problem of linearity in face modeling by using Deep Boltzmann Machines. However, they only work with 2D face and sparse landmarks; and hence cannot handle faces with large-pose variations or occlusion well.

\Paragraph{2D Face Alignment}
2D Face Alignment can be cast as a regression problem where 2D landmark locations are regressed directly~\cite{dollar2010cascaded}. 
For large-pose or occluded faces, strong priors of 3DMM face shape have been shown to be beneficial. 
Hence, there is increasing attention in conducting face alignment by fitting a 3D face model to a single 2D image~\cite{jourabloo2015pose, jourabloo2016large, zhu2016face, liu2016joint, mcdonagh2016joint,  jourabloo2017pose, jourabloo2017poseinvariant}. 
Among the prior works, iterative approaches with cascades of regressors tend to be preferred. 
At each cascade, it can be a single~\cite{tulyakov2015regressing, jourabloo2015pose} or even two regressors~\cite{wu2015robust}. 
In contrast to aforementioned works that use a fixed 3DMM model, our model and model fitting are learned jointly. 
This results in a more powerful model: a single-pass encoder, which is learnt jointly with the model, achieves state-of-the-art face alignment performance on AFLW2000~\cite{zhu2016face} benchmark dataset.  

\begin{figure*}[t!]
\centering
\includegraphics[trim=0 40 0 15,clip, width=0.8\linewidth]{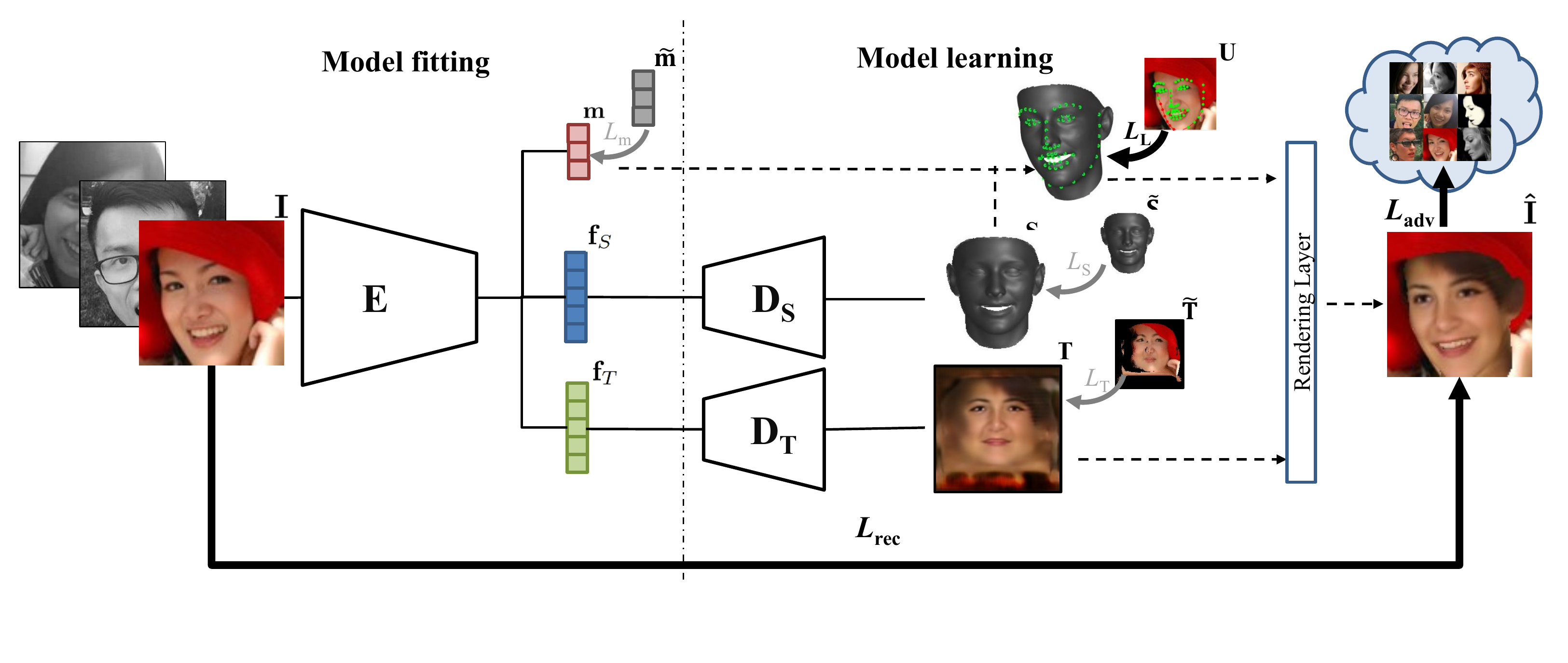}
\vspace{-3mm}
\caption{\small Jointly learning a nonlinear 3DMM and its fitting algorithm from unconstrained 2D face images, in a weakly supervised fashion.}
\label{fig:architecture}
\figvspace 
\end{figure*}

\Paragraph{3D Face Reconstruction}
3DMM also demonstrates its strength in face reconstruction. 
Since with a single image, present information about the surface is limited; 3D face reconstruction must rely on prior knowledge like 3DMM~\cite{adaptive-3d-face-reconstruction-from-unconstrained-photo-collections}. 
Besides 3DMM fitting methods~\cite{blanz2003face, gu2008generative, zhang2006face,tewari2017mofa}, recently, Richardson et al.~\cite{richardson2017learning} design a refinement network that adds facial details on top of the 3DMM-based geometry.
However, this approach can only learn 2.5D depth map, which loses the correspondence property of 3DMM.
The recent work of Tewari et al.~reconstruct a 3D face by an elegant encoder-decoder network~\cite{tewari2017mofa}.
While their ability to decompose lighting with reflectance is satisfactory, our work has a different objective of learning a nonlinear 3DMM.

\Section{Proposed Method}
\label{sec:alg}

\SubSection{Conventional Linear 3DMM}
The 3D Morphable Model (3DMM)~\cite{blanz1999morphable} and its 2D counterpart, Active Appearance Model~\cite{cootes2001active,face-model-fitting-on-low-resolution-images}, provide parametric models for synthesizing faces, where faces are modeled using two components: shape and texture. 
In~\cite{blanz1999morphable}, Blanz et al.~propose to describe the 3D face space with PCA:
\begin{equation}
\mathbf{S} = \mathbf{\bar{S}} +  \mathbf{A} \mathbf{\alpha},
\eqnvspace
\end{equation}
where $\mathbf{S}\in\mathbb{R}^{3Q}$ is a 3D face with $Q$ vertices, $\mathbf{\bar{S}}\in\mathbb{R}^{3\times Q}$ is the mean shape, $\alpha\in\mathbb{R}^{l_S}$ is the shape parameter corresponding to a 3D shape bases $\mathbf{A}$. 
The shape bases can be further split into $\mathbf{A} = [\mathbf{A}_{id}, \mathbf{A}_{exp}]$, where $\mathbf{A}_{id}$ is trained from 3D scans with neutral expression, and $\mathbf{A}_{exp}$ is from the offsets between expression and neutral scans.

The texture $\mathbf{T}^{(l)}\in\mathbb{R}^{3Q}$ of the face is defined within the mean shape $\mathbf{\bar{S}}$, which describes the R, G, B colors of $Q$  corresponding vertices.
$\mathbf{T}^{(l)}$ is also formulated as a linear combination of texture basis functions:
\begin{equation}
\mathbf{T}^{(l)} = \mathbf{\bar{T}}^{(l)} + \mathbf{B} \mathbf{\beta},
\eqnvspace
\end{equation}
where $\mathbf{\bar{T}}^{(l)}$ is the mean texture, $\mathbf{B}$ is the texture bases, and $\mathbf{\beta}\in\mathbb{R}^{l_T}$ is the texture parameter.

The 3DMM can be used to synthesize novel views of the face. 
Firstly, a 3D face is projected onto the image plane with the weak perspective projection model:
\begin{equation}
g(\alpha, \mathbf{m})= \mathbf{V} = f \ast \mathbf{Pr}\ast\mathbf{R}\ast\mathbf{S}+\mathbf{t}_{2d} =  M(\mathbf{m}) \ast \begin{bmatrix} \mathbf{S} \\ \mathbf{1} \end{bmatrix},
\label{eqn:projection}
\end{equation}
where $g(\alpha, \mathbf{m})$ is the model construction and projection function leading to the 2D positions $\mathbf{V}$ of 3D vertices, $f$ is the scale factor, 
$\mathbf{Pr} = \begin{bmatrix} 1 & 0 & 0 \\ 0 & 1 & 0 \end{bmatrix} $ is the orthographic projection matrix,
$\mathbf{R}$ is the rotation matrix constructed from three rotation angles pitch, yaw, roll, and $\mathbf{t}_{2d}$ is the translation vector. 
While the projection matrix $M$ has dimensions $2 \times 4$, it has six degrees of freedom, which is parameterized by a $6$-dim vector $\mathbf{m}$. 
Then, the 2D image is rendered using texture and an illumination model as described in~\cite{blanz1999morphable}.

\SubSection{Nonlinear 3DMM}

As mentioned in Sec.~\ref{sec:intro}, the linear 3DMM has the problems such as requiring 3D face scans for supervised learning, unable to leverage massive unconstrained face images for learning, and the limited representation power due to the linear bases.
We propose to learn a nonlinear 3DMM model using only large-scale in-the-wild 2D face images.

\SubSubSection{Problem Formulation}

In linear 3DMM, the factorization of each components (texture, shape)  can be seen as a matrix multiplication between coefficients and bases. 
From a neural network's perspective, this can be viewed as a shallow network with only {\it one fully connected layer} and no activation function. 
Naturally, to increase the model's representative power, the shallow network can be extended to a deep architecture. 
In this work, we design a novel learning scheme to learn a deep 3DMM and its inference (or fitting) algorithm. 

Specifically, as shown in Fig.~\ref{fig:architecture}, we use two deep networks to decode the shape, texture parameters into the 3D facial shape and texture respectively. 
To make the framework end-to-end trainable, these parameters are estimated by an encoder network, which is essentially the fitting algorithm of our 3DMM.
Three deep networks join forces for the ultimate goal of reconstructing the input face image, with the assistance of a geometry-based rendering layer. 

Formally, given a set of 2D face images $\{\mathbf{I}_i \}_{i=1}^N$, we aim to learn an encoder $E$: $\mathbf{I}{\rightarrow}\mathbf{m},\mathbf{f}_S, \mathbf{f}_T$ that estimates the projection parameter $\mathbf{m}$, and shape and texture parameters $\mathbf{f}_S\in\mathbb{R}^{l_S}, \mathbf{f}_T\in\mathbb{R}^{l_T}$, a 3D shape decoder $D_S$: $\mathbf{f}_S{\rightarrow} \mathbf{S}$ that decodes the shape parameter to a 3D shape $\mathbf{S}$, and a texture decoder $D_T$: $\mathbf{f}_T{\rightarrow} \mathbf{T}$ that decodes the texture parameter to a realistic texture $\mathbf{T}\in\mathbb{R}^{U\times V}$, with the objective that the rendered image with $\mathbf{m}$, $\mathbf{S}$, and $\mathbf{T}$ can approximate the original image well. 
Mathematically, the objective function is:

\noindent \resizebox{0.90\linewidth}{!}{
\begin{minipage}{\linewidth}
\eqnvspace
\vspace{-0.5mm}
\begin{eqnarray}
\argmin_{E,D_S, D_T} \sum_{i=1}^N \norm{\mathcal{R}( E_m(\mathbf{I}_i), D_S(E_S(\mathbf{I}_i)), D_T(E_T(\mathbf{I}_i)) ) -  \mathbf{I}_i}_1,
\end{eqnarray}
\vspace{0.5mm}
\end{minipage}
}
where $\mathcal{R}(\mathbf{m},\mathbf{S},\mathbf{T})$ is the rendering layer (Sec.~\ref{sec:rendering}).

\begin{figure}[t!]
\centering
\includegraphics[trim=0 0 0 0,clip, width=0.85\linewidth]{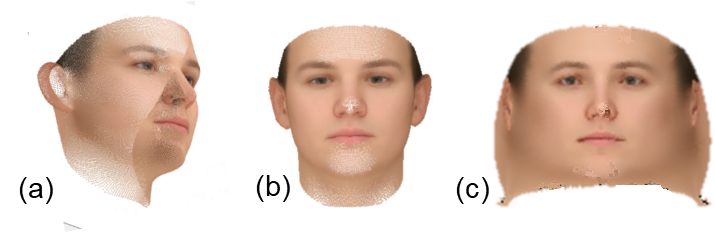}
\vspace{-3mm}
\caption{\small Three texture representations. (a) Texture value per vertex, (b) Texture as a 2D frontal face, (c) 2D unwarped texture.}
\label{fig:tex_representation}
\figvspace 
\end{figure}

\SubSubSection{Shape \& Texture Representation}
Our shape representation is the same as that of the linear 3DMM, i.e., $\mathbf{S}\in\mathbb{R}^{3\times Q}$ is a set of $Q$ vertices $ \mathbf{v}_S = (x,y,z)$ on the face surface. 
The shape decoder $D_S$ is a MLP whose input is the shape parameter $\mathbf{f}_S$ from $E$.

Fig.~\ref{fig:tex_representation} illustrates three possible texture representations. 
Texture is defined per vertex in the linear 3DMM and recent work such as~\cite{tewari2017mofa} (Fig.~\ref{fig:tex_representation}(a)). 
There is a texture intensity value corresponding to each vertex in the face mesh. 
Since 3D vertices are not defined on a 2D grid, this representation will be parameterized as a vector, which not only loses the spatial relation of vertices, but also prevents it from leveraging the convenience of deploying CNN on 2D imagery.
In contrast, given the rapid progress in image synthesis, it is desirable to choose a 2D image, e.g., a frontal-view face image in Fig.~\ref{fig:tex_representation}(b), as a texture representation. 
However, frontal faces contain little information of two sides, which would lose much texture information for side-view faces.

In light of these considerations, we use an unwrapped 2D texture as our texture representation (Fig.~\ref{fig:tex_representation}(c)).
Specifically, each 3D vertex $\mathbf{v}_S$  is projected onto the UV space using cylindrical unwarp. 
Assuming that the face mesh has the top pointing up the $y$ axis, the projection of $\mathbf{v}_S = (x, y, z)$ onto the UV space $\mathbf{v}_T = (u, v)$ is computed as:
\begin{equation}
 v \rightarrow \alpha_1 . \text{arctan} \left( \frac{x}{z} \right) + \beta_1, \hspace{3mm} u \rightarrow \alpha_2 . y + \beta_2,
 \label{eqn:unwarp}
\end{equation}
where $\alpha_1, \alpha_2, \beta_1, \beta_2$ are constant scale and translation scalars to place the unwrapped face into the image boundaries.
Also, the texture decoder $D_T$ is a CNN constructed by fractionally-strided convolution layers.

\SubSubSection{In-Network Face Rendering}
\label{sec:rendering}

\begin{figure}[t!]
\centering
\includegraphics[trim=39 30 35 0,clip, width=0.9\linewidth]{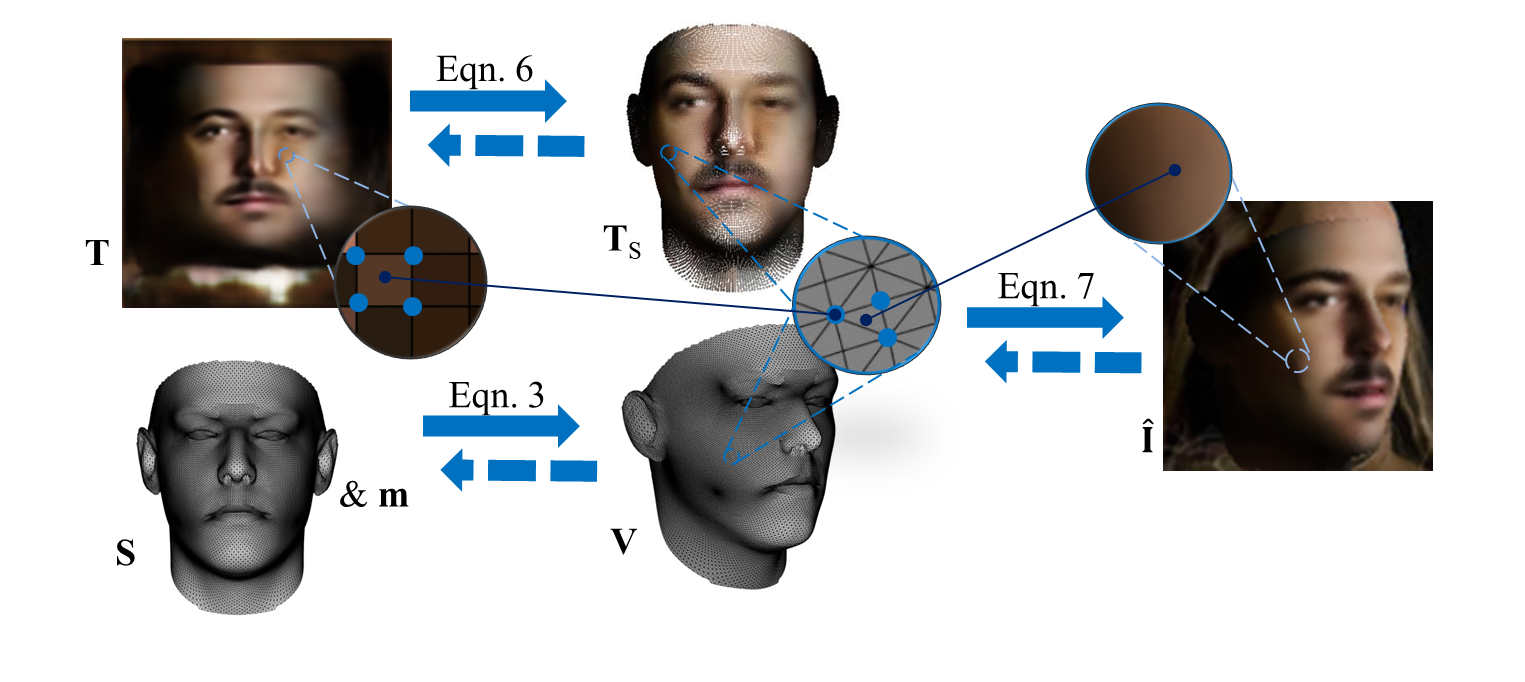}
\vspace{-3mm}
\caption{\small Forward and backward pass of the rendering layer.}
\label{fig:structure}
\figvspace 
\end{figure}

To reconstruct a face image from the texture $\mathbf{T}$, shape $\mathbf{S}$, and projection parameter $\mathbf{m}$, we define a rendering layer $\mathcal{R}(\mathbf{m},\mathbf{S},\mathbf{T})$. 
This is accomplished in three steps. 
Firstly, the texture value of each vertex in $\mathbf{S}$ is determined by its predefined location in the 2D texture $\mathbf{T}$. 
Usually, it involves sub-pixel sampling via a bilinear sampling kernel:
\noindent \resizebox{ 0.98\linewidth}{!}{
\begin{minipage}{\linewidth}
\begin{eqnarray}
\mathbf{T}_S(\mathbf{v}_S) = \! \! \! \sum_{ \substack{ u' \in \{ \floor{u}, \ceil{u} \} \\ v' \in \{ \floor{v}, \ceil{v} \} }} \! \! \! \mathbf{T}(u',v')(1{-}|u{-}u'|) (1{-}|v{-}v'|),
\end{eqnarray}
\end{minipage}
}
where $\mathbf{v}_T = (u, v)$ is the UV space projection of $\mathbf{v}_S$ via Eqn.~\ref{eqn:unwarp}.
Secondly, the 3D shape/mesh $\mathbf{S}$ is projected to the image plane via Eqn.~\ref{eqn:projection}.
Finally, the 3D mesh is then rendered using a Z-buffer renderer, where each pixel is associated with a single triangle of the mesh,
\begin{align}
\mathbf{\hat{I}}(m, n) & = \mathcal{R}(\mathbf{m},\mathbf{S}, \mathbf{T})_{m,n}  &= \! \! \! \sum_{\mathbf{v}_S \in \Phi(g, m, n) } \! \! \! \lambda \mathbf{T}_S(\mathbf{v}_S), 
\end{align}
where $\Phi(g, m, n) = \{\mathbf{v}_S ^{(1)}, \mathbf{v}_S ^{(2)}, \mathbf{v}_S^{(3)} \}$ is an operation returning three vertices of the triangle that encloses the pixel $(m, n)$ after projection $g$. 
In order to handle occlusions, when a single pixel resides in more than one triangle, the triangle that is closest to the image plane is selected. 
The value of each pixel is determined by interpolating the intensity of the mesh vertices via barycentric coordinates $\{\lambda^{(i)}\}_{i=1}^3$.

There are alternative designs to our rendering layer. 
If the texture representation is defined per vertex, as in Fig.~\ref{fig:tex_representation}(a), one may warp the input image $\mathbf{I}_i$ onto the vertex space of the 3D shape $\mathbf{S}$, whose distance to the per-vertex texture representation can form a reconstruction loss. 
This design is adopted by the recent work of~\cite{tewari2017mofa}.
In comparison, our rendered image is defined on a 2D grid while the alternative is on top of the 3D mesh.
As a result, our rendered image can enjoy the convenience of applying the adversarial loss, which is shown to be critical in improving the quality of synthetic texture.
%
Another design for rendering layer is image warping based on the spline interpolation, as in~\cite{cole2017face}. 
However, this warping is continuous: every pixel in the input will map to the output. 
Hence this warping operation fails in the occlusion part. 
As a result, Cole et al.~\cite{cole2017face} limit their scope to only synthesizing frontal faces by warping from normalized faces. \footnote{Our rendering layer implementation is publicly available at~\url{https://github.com/tranluan/Nonlinear_Face_3DMM}.}

\begin{table}[t!]
\caption{\small The structures of $E$ and $D_T$ networks}
\vspace{-6mm}
\label{tab:network}
\begin{center}
\small
\resizebox{0.99\linewidth}{!}{
\setlength{\tabcolsep}{3pt}
\begin{tabular}{ @{}ccccccccc@{} }
\toprule
\multicolumn{3}{c}{$E$} & \hspace{2mm} 
& \multicolumn{3}{c}{$D_T$} \\
\cmidrule(r){1-3}
\cmidrule(r){5-7}
Layer & Filter/Stride & Output Size && Layer & Filter/Stride & Output Size \\ \midrule
&&&& FC & & $8{\times}8{\times}320$ \\
Conv11 & $3{\times}3/1$ & $96{\times}96{\times}32$ && FConv52& $3{\times}3/1$ & $8{\times}8{\times}160$ \\
Conv12 & $3{\times}3/1$ & $96{\times}96{\times}64$ && FConv51& $3{\times}3/1$ & $8{\times}8{\times}256$ \\\midrule
Conv21 & $3{\times}3/2$ & $48{\times}48{\times}64$ && FConv43& $3{\times}3/2$ & $16{\times}16{\times}256$&&\\ 
Conv22 & $3{\times}3/1$ & $48{\times}48{\times}64$  && FConv42& $3{\times}3/1$ & $16{\times}16{\times}128$ \\
Conv23 & $3{\times}3/1$ & $48{\times}48{\times}128$ && FConv41& $3{\times}3/1$ & $16{\times}16{\times}192$ \\
\midrule
Conv31 & $3{\times}3/2$ & $24{\times}24{\times}128$ && FConv33& $3{\times}3/2$ & $32{\times}32{\times}192$ && \\ 
Conv32 & $3{\times}3/1$ & $24{\times}24{\times}96$  && FConv32& $3{\times}3/1$ & $32{\times}32{\times}96$ \\
Conv33 & $3{\times}3/1$ & $24{\times}24{\times}192$ && FConv31& $3{\times}3/1$ & $32{\times}32{\times}128$ \\
\midrule
Conv41 & $3{\times}3/2$ & $12{\times}12{\times}192$ && FConv23& $3{\times}3/2$ & $64{\times}64{\times}128$ \\ 
Conv42 & $3{\times}3/1$ & $12{\times}12{\times}128$ && FConv22& $3{\times}3/1$ & $64{\times}64{\times}64$ \\
Conv43 & $3{\times}3/1$ & $12{\times}12{\times}256$ && FConv21& $3{\times}3/1$ & $64{\times}64{\times}64$ \\
\midrule
Conv51 & $3{\times}3/2$ & $6{\times}6{\times}256$ && 
FConv13& $3{\times}3/2$ & $128{\times}128{\times}64$ \\ 
Conv52 & $3{\times}3/1$ & $6{\times}6{\times}160$ && FConv12& $3{\times}3/1$ & $128{\times}128{\times}32$ \\
Conv53 & $3{\times}3/1$ & $6{\times}6{\times}(l_S{+}l_T{+}64)$ && FConv11& $3{\times}3/1$ & $128{\times}128{\times}3$ \\
\midrule
AvgPool& $6{\times}6/1$ & $1{\times}1{\times}(l_S{+}l_T{+}64)$ && \\ \midrule
FC (for $\mathbf{m}$ only) & $64{\times}6$& $6$ \\ \bottomrule
\end{tabular}}
\vspace{-6mm}
\end{center}
\end{table}

\SubSubSection{Network Architecture}
We design our $E, D_T$ network architecture as in Tab.~\ref{tab:network}. Also, $D_S$ includes two fully connected layers with \mbox{$1,000$-dim} intermediate representation with eLU activation.

The entire network is end-to-end trained to reconstruct the input images, with the loss function:
\begin{equation}
L = L_{\text{rec}} + \lambda_{ \text{adv} } L_{ \text{adv} } +  \lambda_L L_{\text{L}},
\label{tab:overallLoss}
\end{equation}
where the reconstruction loss $L_{\text{rec}} = \sum_{i=1}^{N}||\mathbf{\hat{I}}_i - \mathbf{I}_i||_1$ enforces the rendered image $\mathbf{\hat{I}}_i$ to be similar to the input $\mathbf{I}_i$, the adversarial loss $L_{ \text{adv} }$ favors realistic rendering, and the landmark loss $L_L$ enforces geometry constraint.

\Paragraph{Adversarial Loss}
Based on the principal of Generative Adversarial Network (GAN)~\cite{goodfellow2014generative}, the adversarial loss is widely used to synthesize photo-realistic images~\cite{radford2015unsupervised, tran2017disentangled, tran2018representation}, where the generator and discriminator are trained alternatively.
In our case, networks that generate the rendered image $\mathbf{\hat{I}}_i$ is the generator.
The discriminator includes a dedicated network $D_A$, which aims to distinguish between the real face image $\mathbf{I}_i$ and rendered image $\mathbf{\hat{I}}_i$.
During the training of the generator, the texture model $D_T$ will be updated with the objective that $\mathbf{\hat{I}}_i$ is being classified as real faces by $D_A$.
Since our face rendering already creates correct global structure of the face image, the global image-based adversarial loss may not be effective in producing high-quality textures on local facial regions.
Therefore, we employ patchGAN~\cite{shrivastava2017learning} in our discriminator.
Here, $D_A$ is a CNN consisting of four $3\times3$ conv layers with stride of $2$, and number of filters are $32$, $64$, $128$ and $1$, respectively. 
Finally, one of key reasons we are able to employ adversarial loss is that we are rendering in the 2D image space, rather than the 3D vertices space or unwrapped texture space.
This shows the necessity and importance of our rendering layer. 
 
\Paragraph{Semi-Supervised Pre-Training}  
Fully unsupervised training using only the mentioned reconstruction and adversarial loss on the rendered image could lead to a degenerate solution, since the initial estimation is far from ideal to render meaningful images. 
Hence, we introduce pre-training loss functions to guide the training in the early iterations.

With face profiling technique, Zhu et al.~\cite{zhu2016face} expands the 300W dataset~\cite{sagonas2016300} into $122,450$ images with the fitted 3DMM shape $\widetilde{\mathbf{S}}$ and projection parameters $\widetilde{\mathbf{m}}$. 
Given $\widetilde{\mathbf{S}}$ and $\widetilde{\mathbf{m}}$, we create the pseudo groundtruth texture $\widetilde{\mathbf{T}}$ by referring every pixel in the UV space back to the input image, i.e., backward of our rendering layer. 
With $\widetilde{\mathbf{m}}$, $\widetilde{\mathbf{S}}$, $\widetilde{\mathbf{T}}$, we define our pre-training loss by: 
\begin{equation}
L_0 = L_{\text{S}} + \lambda_T L_{\text{T}} + \lambda_m L_{\text{m}} + \lambda_L L_{\text{L}},
\end{equation}
where
\begin{align}
L_{\text{S}} = & \norm{ \mathbf{S}-\mathbf{\widetilde{S}} }_2, \\
L_{\text{T}} = & \norm{ \mathbf{T}-\mathbf{\widetilde{T}} }_1, \\
L_{\text{m}} = & \norm{ \mathbf{m}-\mathbf{\widetilde{m}} }_2. 
\end{align}
Due to the pseudo groundtruth, using $L_0$ may run into the risk that our solution learns to mimic the linear model. 
Thus, we switch to the loss of Eqn.~\ref{tab:overallLoss} after $L_0$ converges.

\Paragraph{Sparse Landmark Alignment}
To help $D_T$ to better learn the facial shape, the landmark loss can be an auxiliary task. 
\begin{align}
L_{\text{L}} = & \norm {M(\mathbf{m}) \ast \begin{bmatrix} \mathbf{S}(:,\mathbf{d}) \\ \mathbf{1} \end{bmatrix} - \mathbf{U} }_2,
\label{eq:landmarkloss}
\end{align}
where $\mathbf{U} \in \mathbb{R}^{2{\times} 68}$ is the manually labeled 2D landmark locations, $\mathbf{d}$ is a constant $68$-dim vector storing the indexes of $68$ 3D vertices corresponding to the labeled 2D landmarks.
Unlike the three losses above, these landmark annotations are ``golden" groundtruth, and hence $L_{\text{L}}$ can be used during the entire training process. 
Different from traditional face alignment work where the shape bases are fixed, our work jointly learns
the bases functions (i.e., the shape decoder $D_S$) as well. 
Minimizing the landmark loss when updating $D_S$ only moves a tiny subset of vertices, since our $D_S$ is a MLP consisting of fully connected layers.
This could lead to unrealistic shapes. 	
Hence, when optimizing the landmark loss, we fix the decoder $D_S$ and only update the encoder.
Note that the estimated groundtruth in $L_0$ and the landmarks are the only supervision used in our training, due to this our learning is considered as {\it weakly} supervised. 

\Section{Experimental Results}
\label{sec:exp}

The experiments study three aspects of the proposed nonlinear 3DMM, in terms of its expressiveness, representation power, and applications to facial analysis.
Using facial mesh triangle definition by Basel Face Model~(BFM)~\cite{paysan20093d}, we train our 3DMM using 300W-LP dataset~\cite{zhu2016face}. 
The model is optimized using Adam optimizer with an initial learning rate of $0.001$ when minimizing $L_0$, and $0.0002$ when minimizing $L$.
We set the following parameters: $Q=53,215$, $U=V=128$, $l_S=l_T=160$. $\lambda$~values are set to make losses to have similar magnitudes.

\begin{figure}[t!]

\begin{center}
\small
\setlength{\tabcolsep}{3pt}
  \resizebox{0.8\linewidth}{!}{
\begin{tabular}{ @{}c@{}c@{}c@{}c@{}c@{}c@{}c@{}c@{}}
\includegraphics[trim=400 180 370 230,clip,width=\VaryingShapeFigWid]{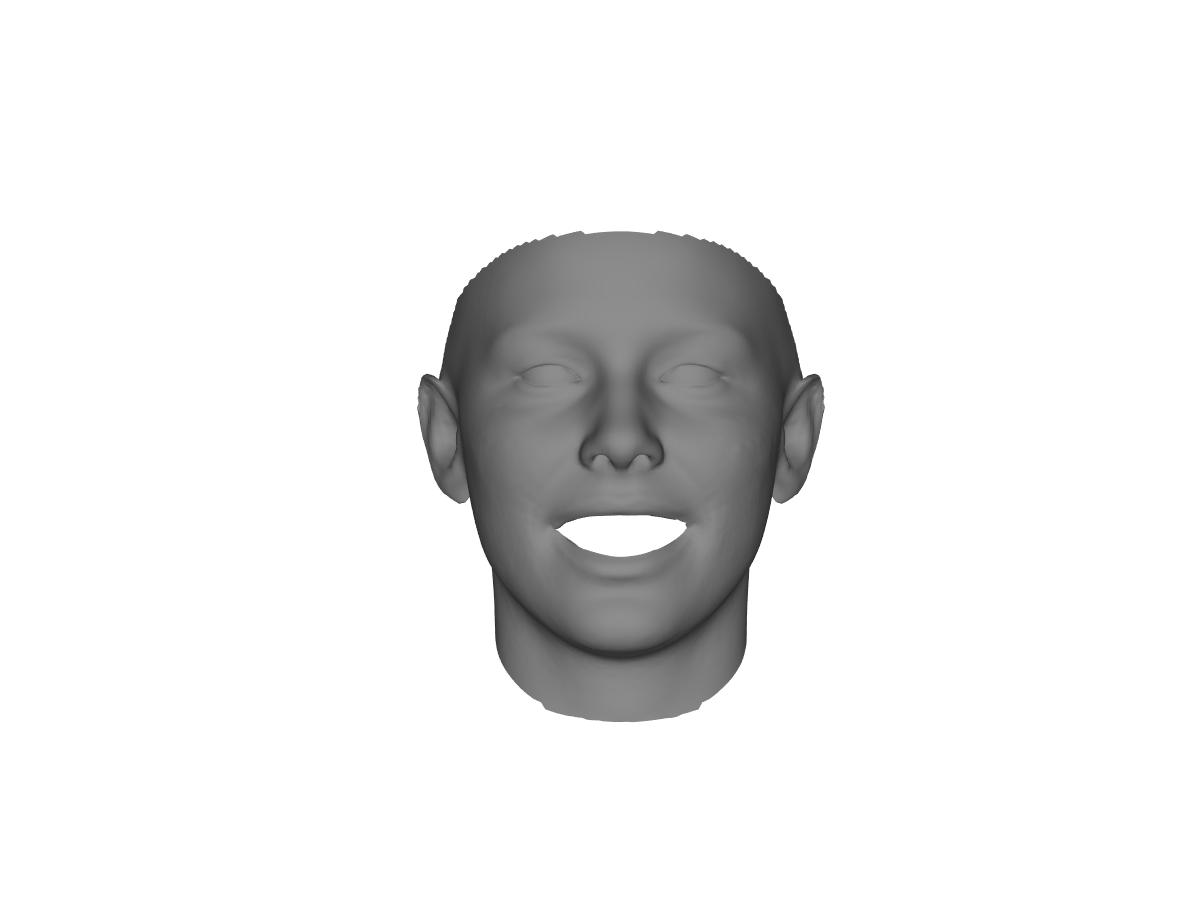} &
\includegraphics[trim=400 180 370 230,clip,width=\VaryingShapeFigWid]{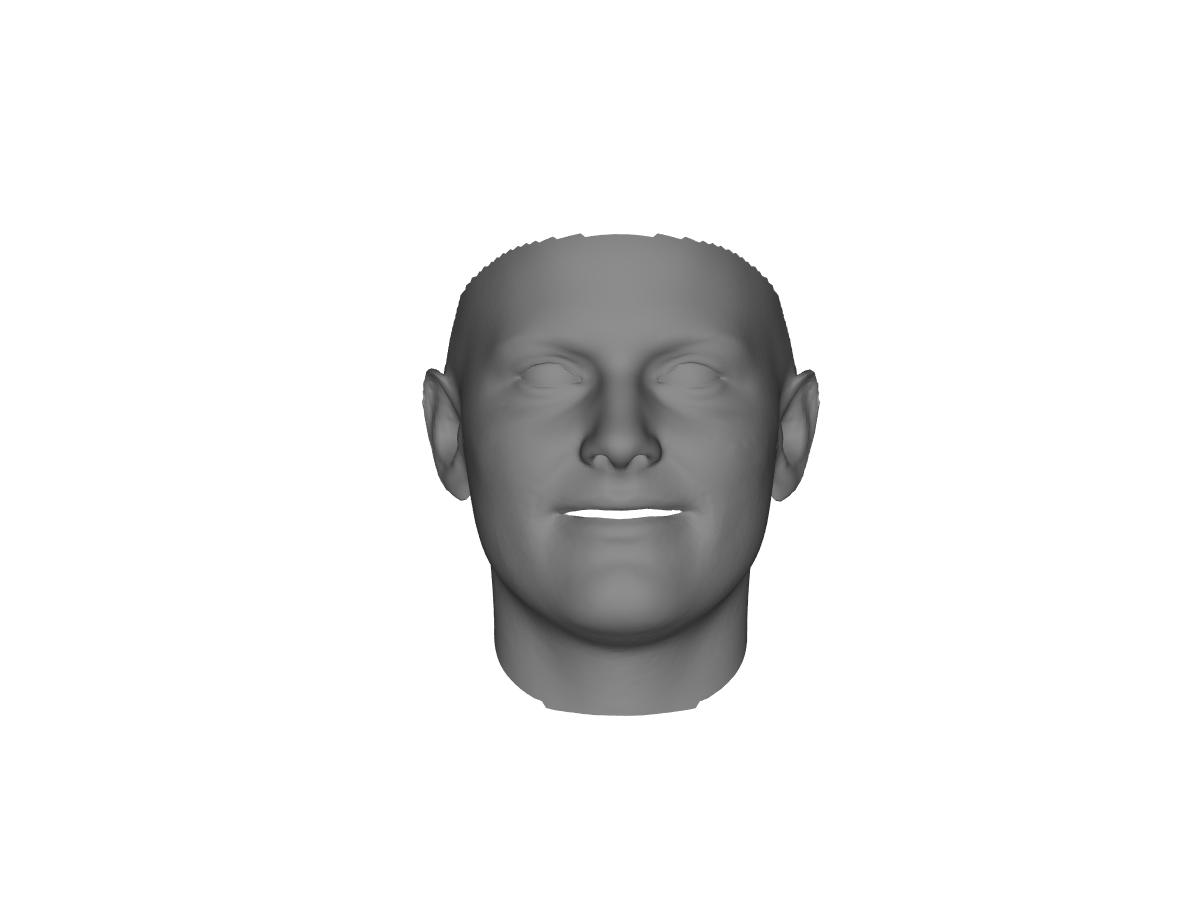} &
\includegraphics[trim=400 180 370 230,clip,width=\VaryingShapeFigWid]{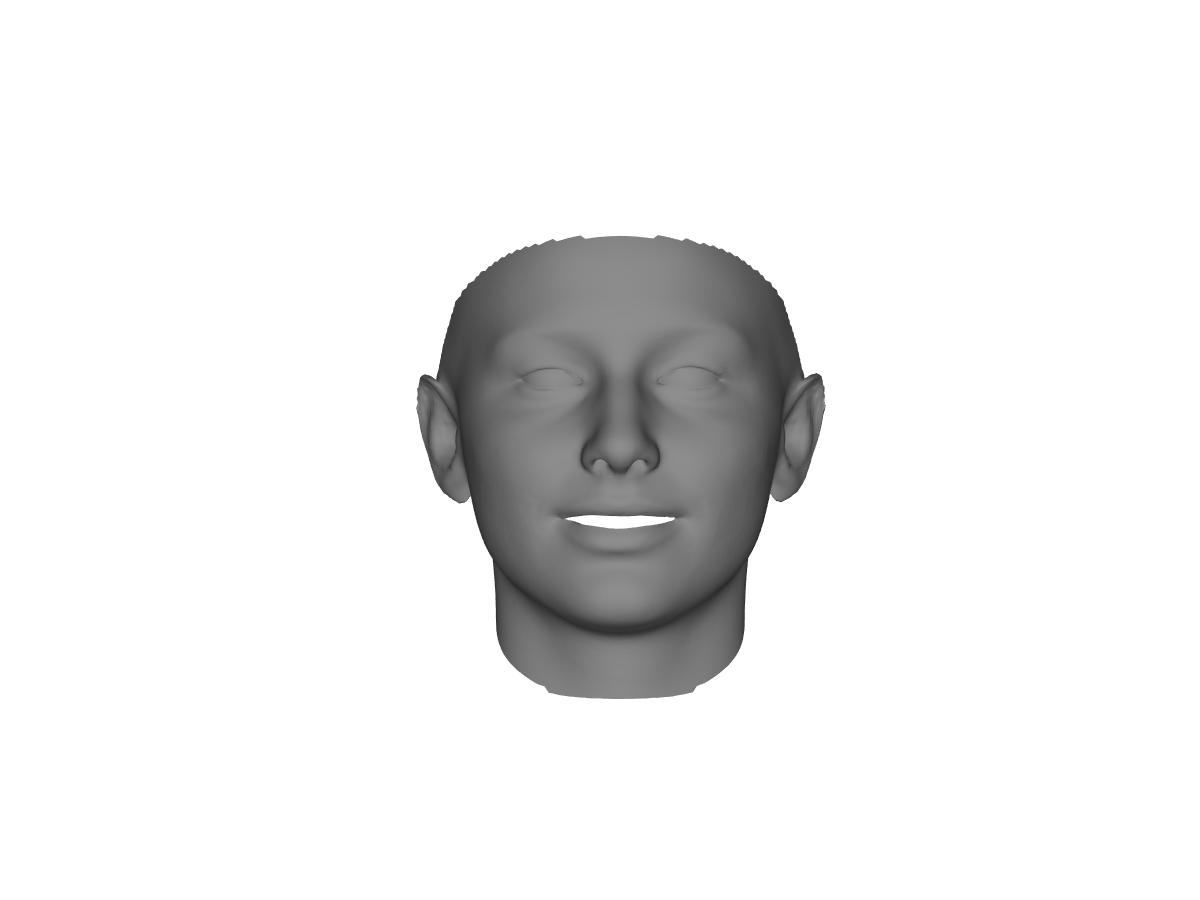} &
\includegraphics[trim=400 180 370 230,clip,width=\VaryingShapeFigWid]{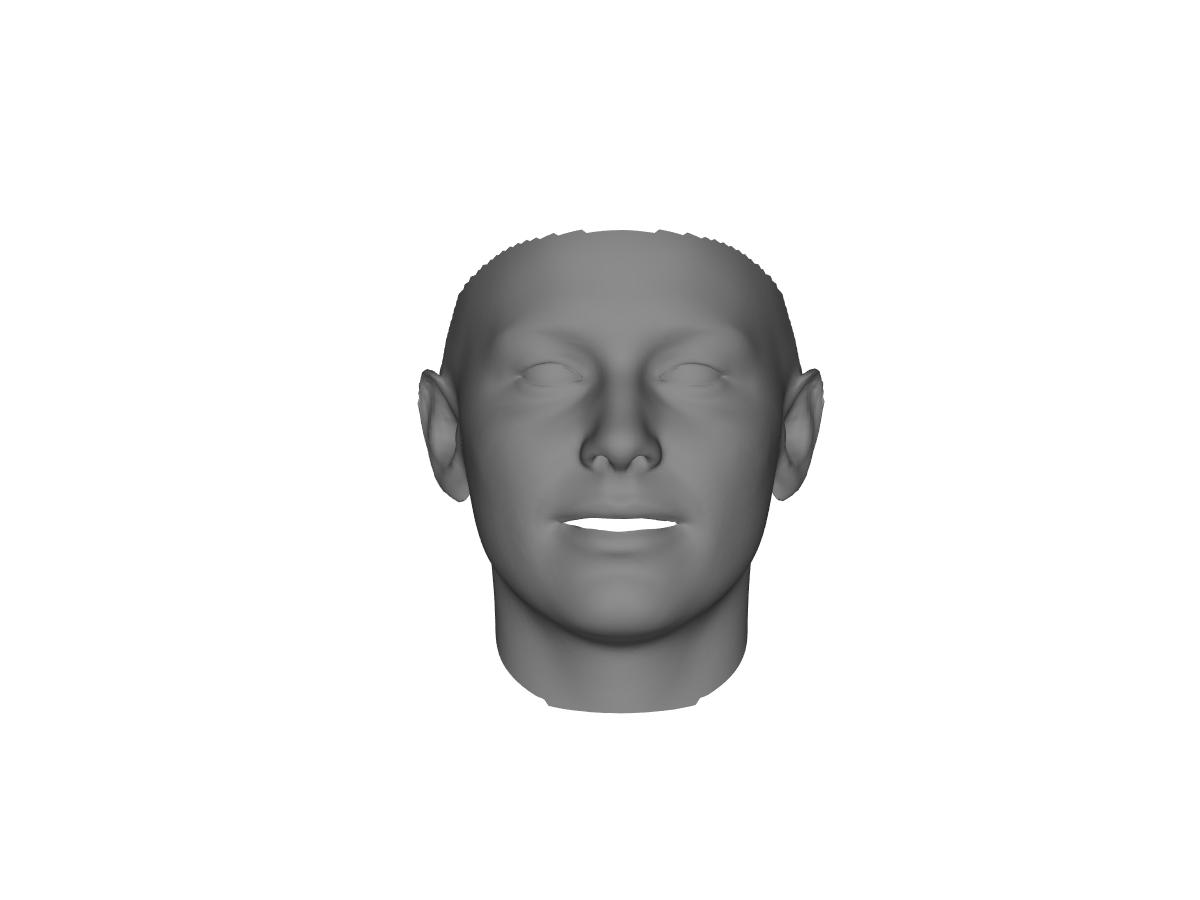} &
\\
\includegraphics[trim=400 180 370 230,clip,width=\VaryingShapeFigWid]{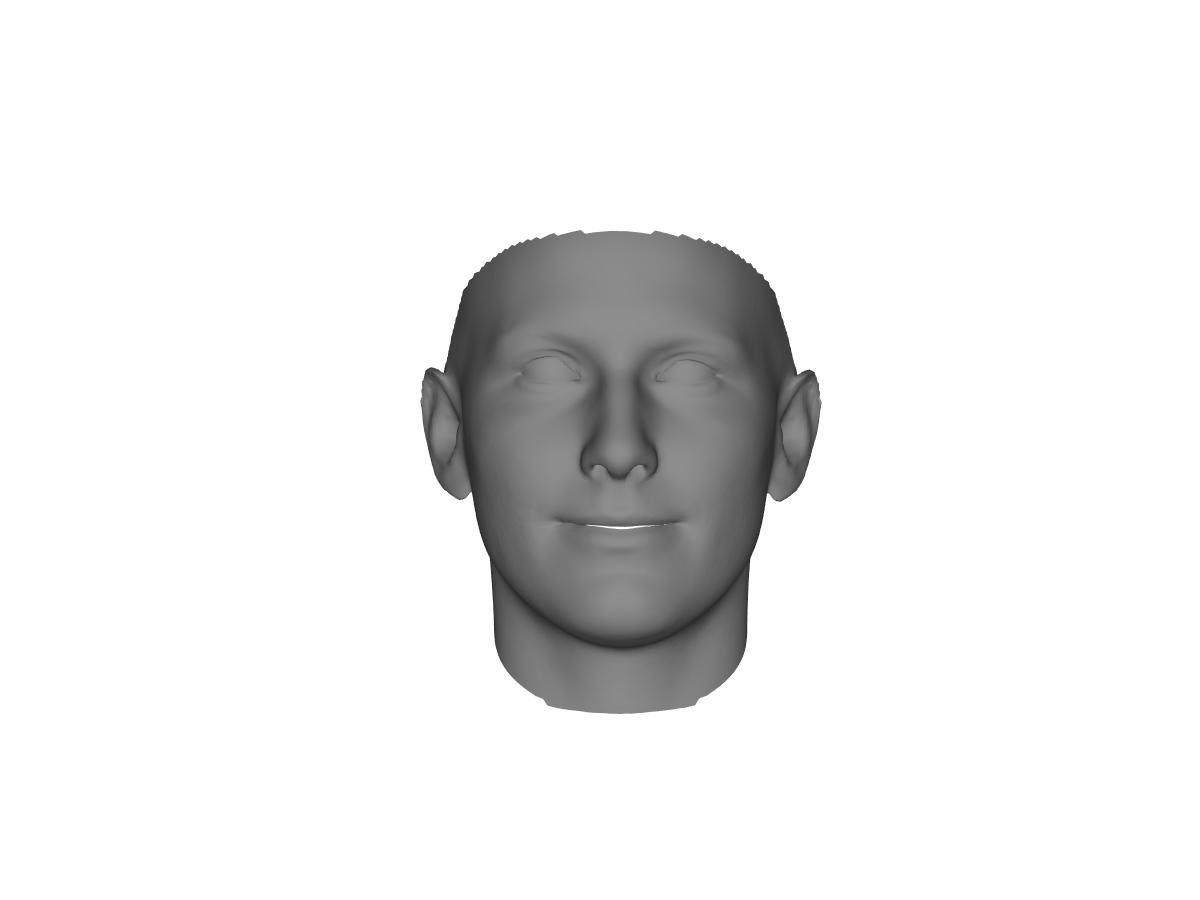} &
\includegraphics[trim=400 180 370 230,clip,width=\VaryingShapeFigWid]{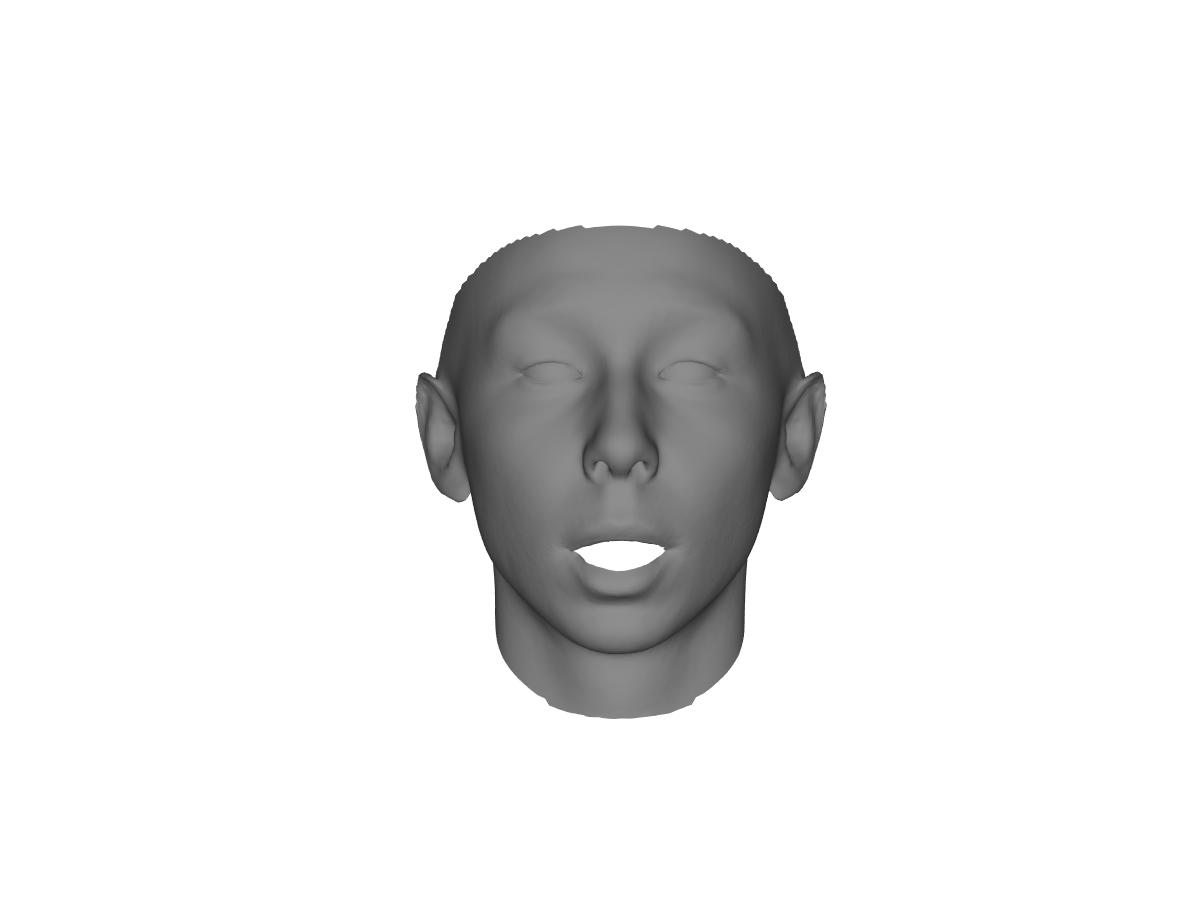} &
\includegraphics[trim=400 180 370 230,clip,width=\VaryingShapeFigWid]{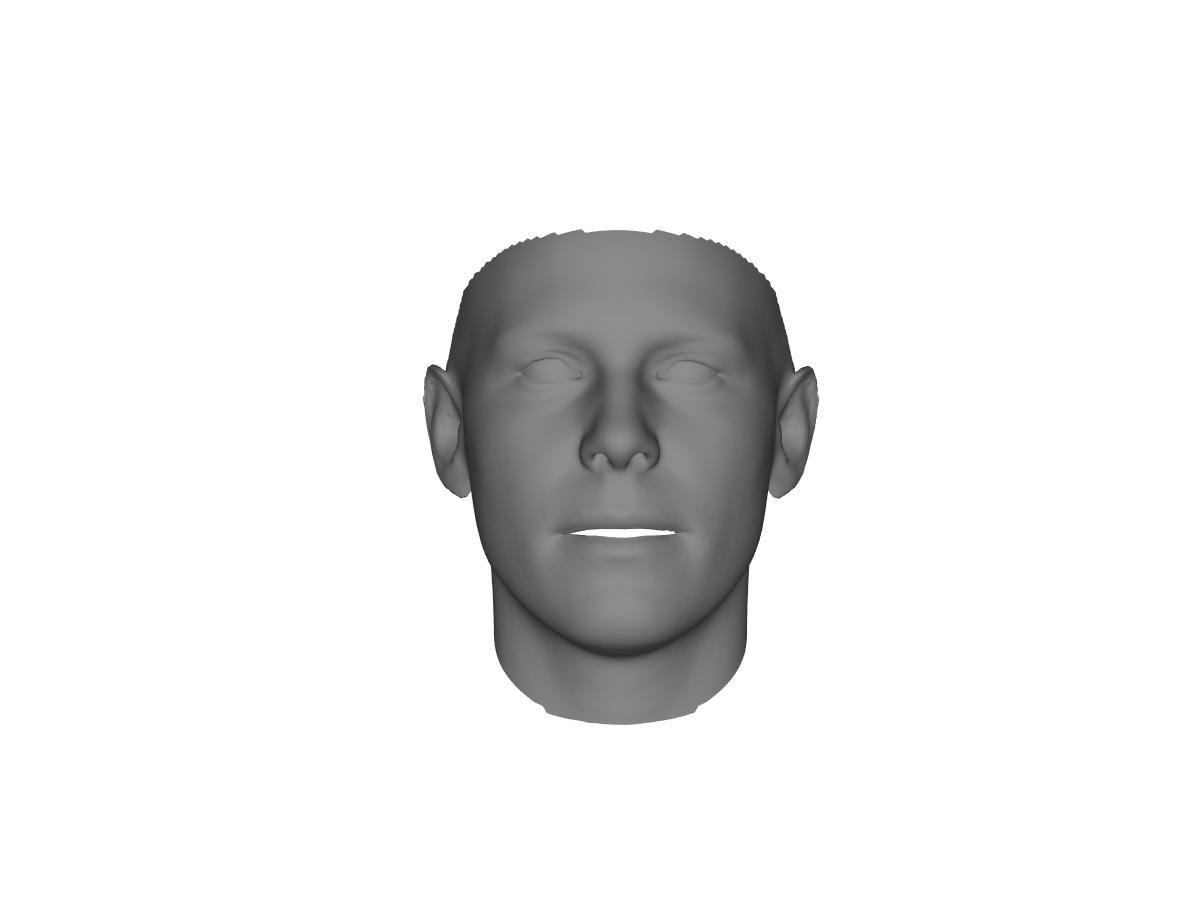} &
\includegraphics[trim=400 180 370 230,clip,width=\VaryingShapeFigWid]{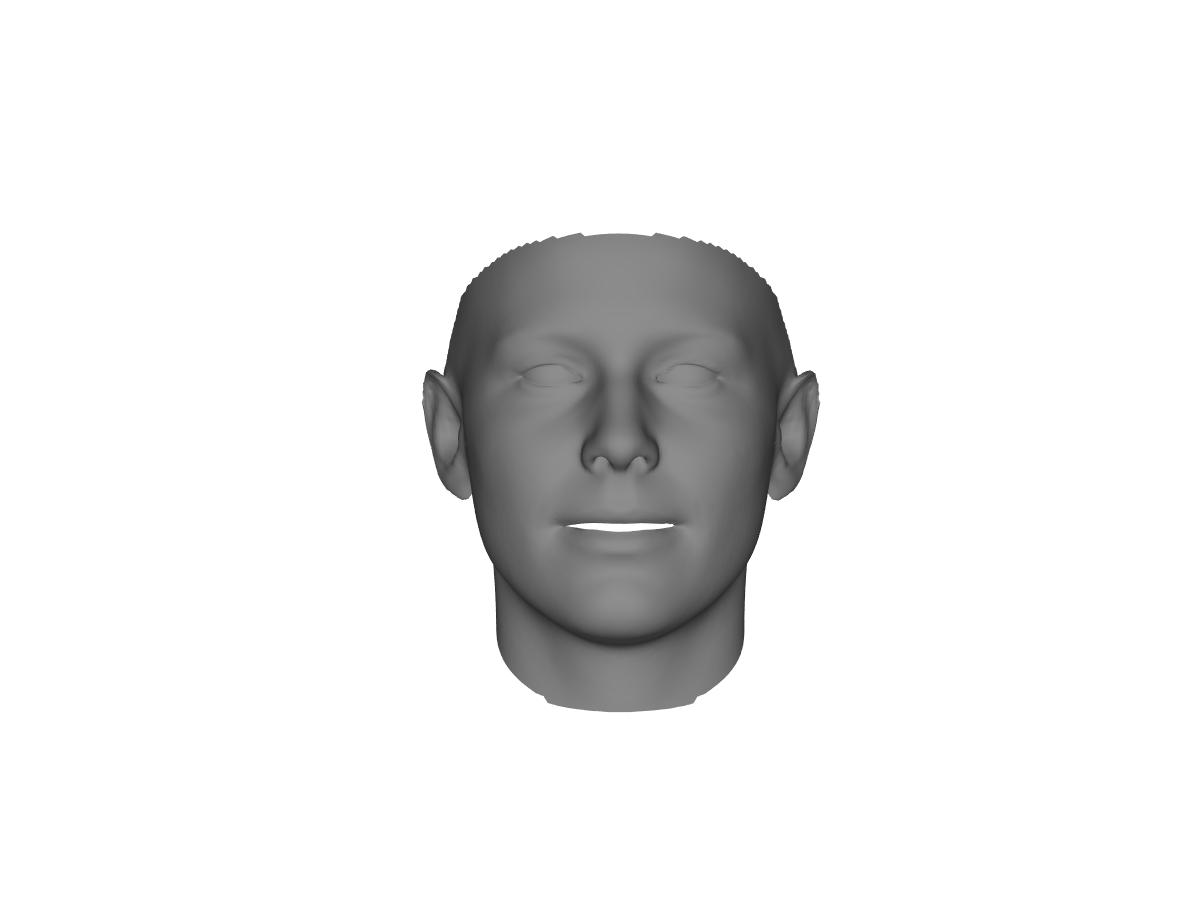} &
\end{tabular}
}
\vspace{-2mm}
\caption{\small Each column shows shape changes when varying one element of $\mathbf{f}_S$. Ordered by the magnitude of shape changes.}
\label{fig:varying_shape}\figvspace \vspace{-2mm}
\end{center}

\end{figure}

\begin{figure}[t!]
\begin{center}
\small
\setlength{\tabcolsep}{3pt}
  \resizebox{0.8\linewidth}{!}{
\begin{tabular}{ @{}c@{}c@{}c@{}c@{}c@{}c@{}c@{}c@{}}
\includegraphics[width=\VaryingShapeFigWid]{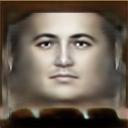} &
\includegraphics[width=\VaryingShapeFigWid]{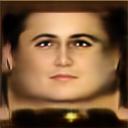} &
\includegraphics[width=\VaryingShapeFigWid]{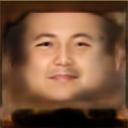} &
\includegraphics[width=\VaryingShapeFigWid]{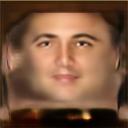} &
\\
\includegraphics[width=\VaryingShapeFigWid]{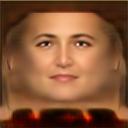} &
\includegraphics[width=\VaryingShapeFigWid]{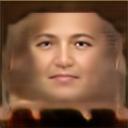} &
\includegraphics[width=\VaryingShapeFigWid]{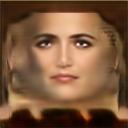} &
\includegraphics[width=\VaryingShapeFigWid]{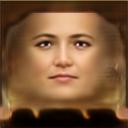} &
\end{tabular}
}
\vspace{-2mm}
\caption{\small Each column shows texture changes when varying one element of $\mathbf{f}_T$.}
\label{fig:varying_tex}\figvspace \vspace{-2mm}
\end{center}
\end{figure}

\SubSection{Expressiveness}
\Paragraph{Exploring feature space}
We use the entire CelebA dataset~\cite{liu2015faceattributes} with ${\sim}200$k images to feed to our network to obtain the empirical distribution of our shape and texture parameters. 
By varying the mean parameter along each dimension proportional to their standard deviations, we can get a sense how each element contributes to the final shape and texture.
We sort elements in the shape parameter $\mathbf{f}_S$ based on their differences to the mean 3D shape. 
Fig.~\ref{fig:varying_shape} shows four examples of shape changes, whose differences rank No.$1$, $40$, $80$, and $120$ among $160$ elements.
Most of top changes are expression related.
Similarly, in Fig.~\ref{fig:varying_tex}, we visualize different texture changes by adjusting only one element of $\mathbf{f}_T$ off the mean parameter $\bar{\mathbf{f}}_T$.
The elements with the same $4$ ranks as the shape counterpart are selected.

\begin{figure}[t!]
\begin{center}
\small
\setlength{\tabcolsep}{3pt}
\begin{tabular}{ @{\hskip 1.5mm}c@{}c@{\hskip 1.5mm}c@{}c@{}c@{}c@{}c@{\hskip 1.5mm}c@{}}
\multicolumn{2}{c}{Male} & \multicolumn{2}{c}{Mustache} & \multicolumn{2}{c}{Pale skin} \\
\includegraphics[width=\afifthcolumn]{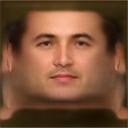} &
\includegraphics[trim=150 250 150 250,clip, width=\afifthcolumn]{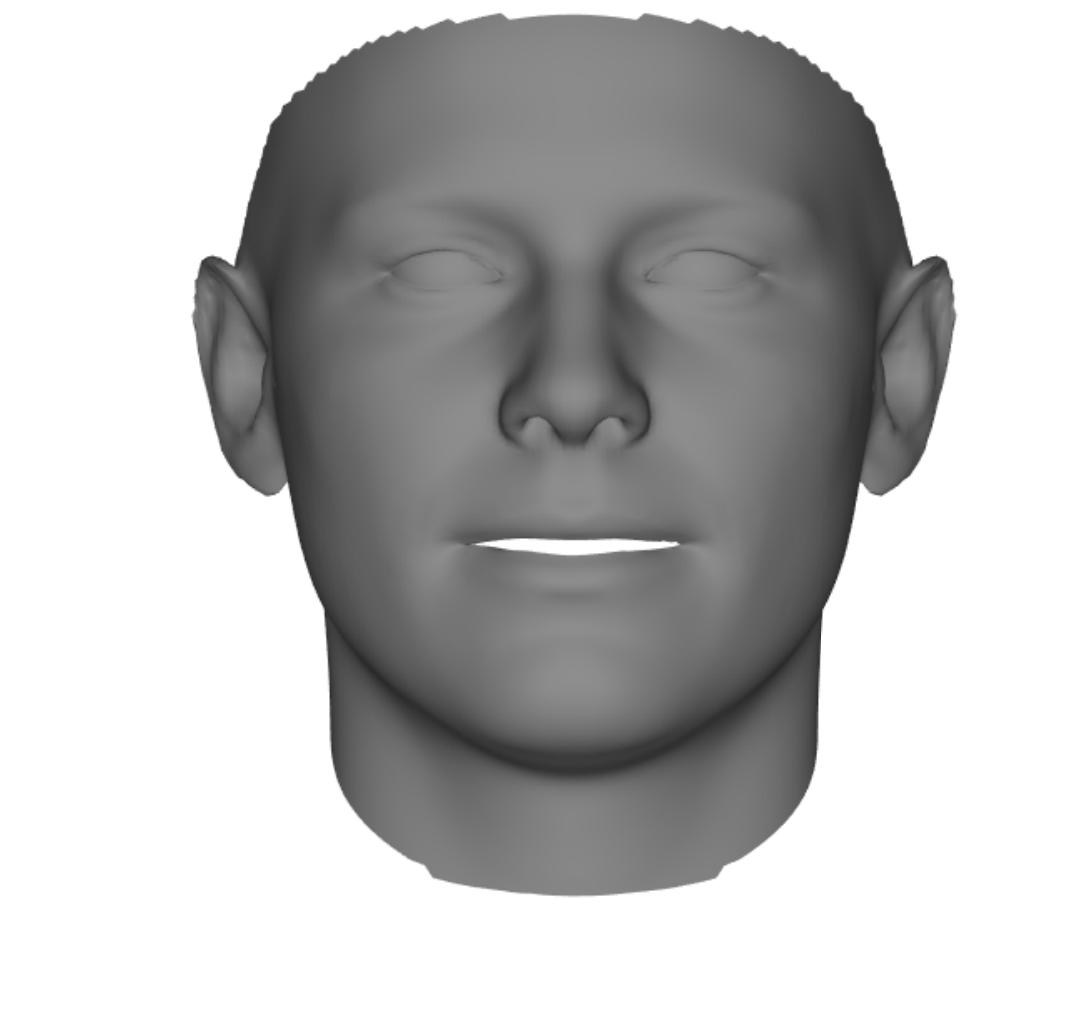} &
\includegraphics[width=\afifthcolumn]{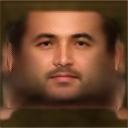} &
\includegraphics[trim=150 250 150 250,clip, width=\afifthcolumn]{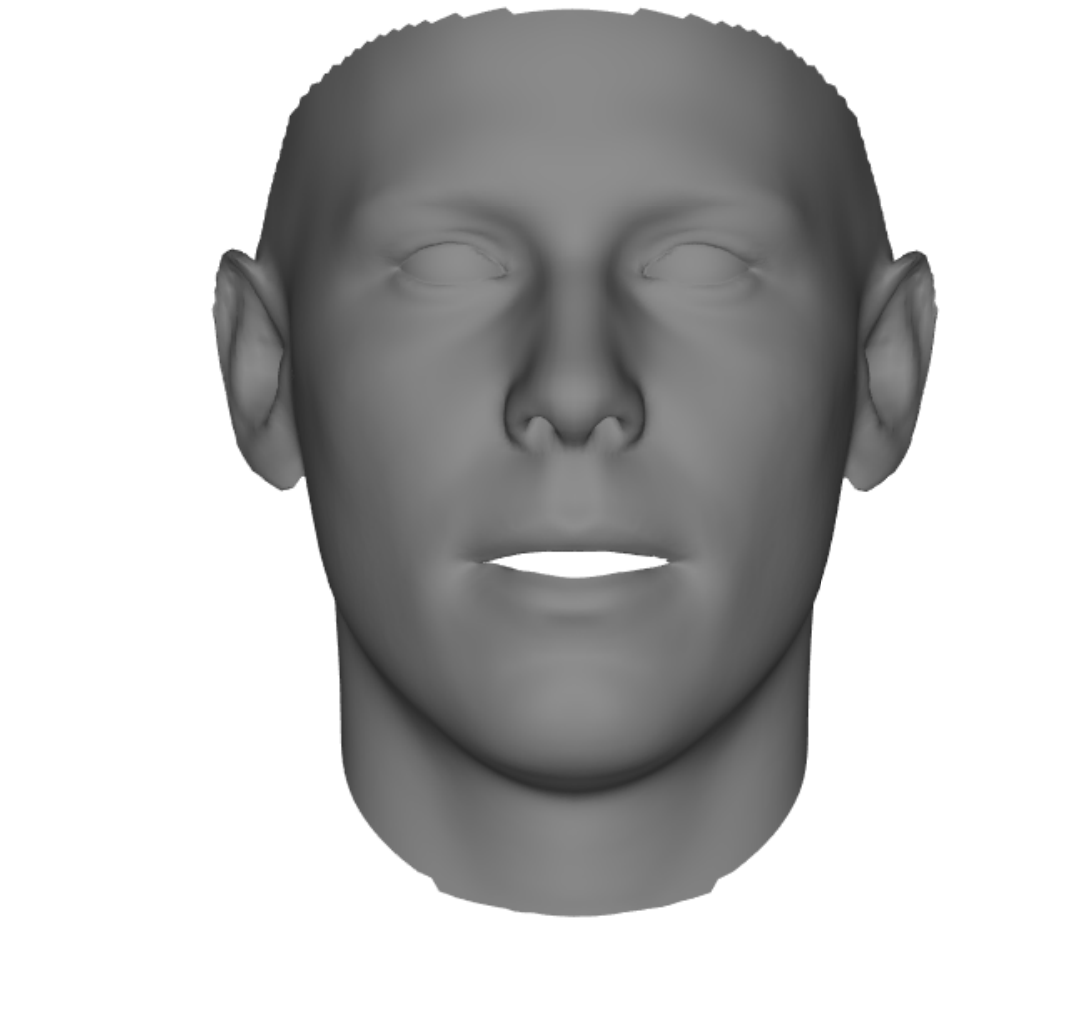} &
\includegraphics[width=\afifthcolumn]{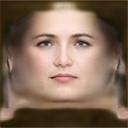} &
\includegraphics[trim=150 250 175 250,clip, width=\afifthcolumn]{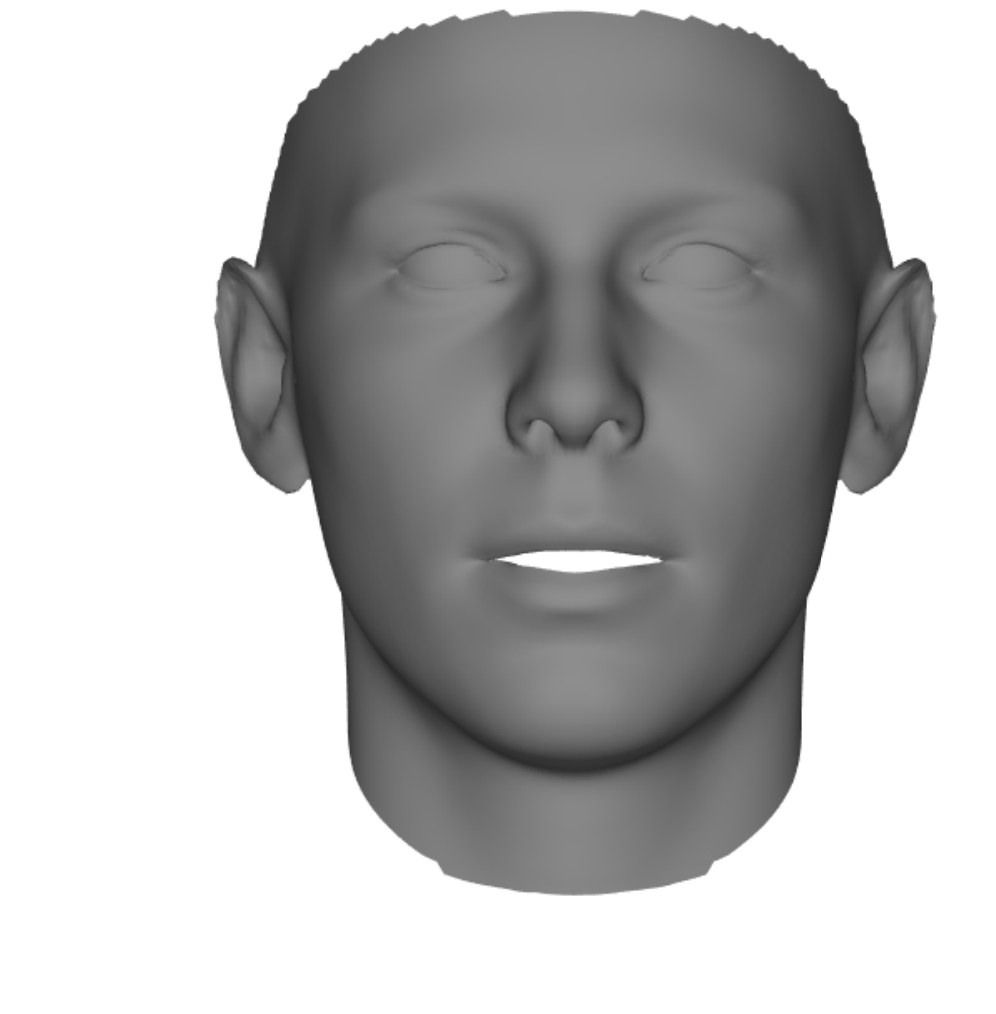} &
\\ 
\includegraphics[width=\afifthcolumn]{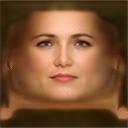} &
\includegraphics[trim=150 250 150 250,clip, width=\afifthcolumn]{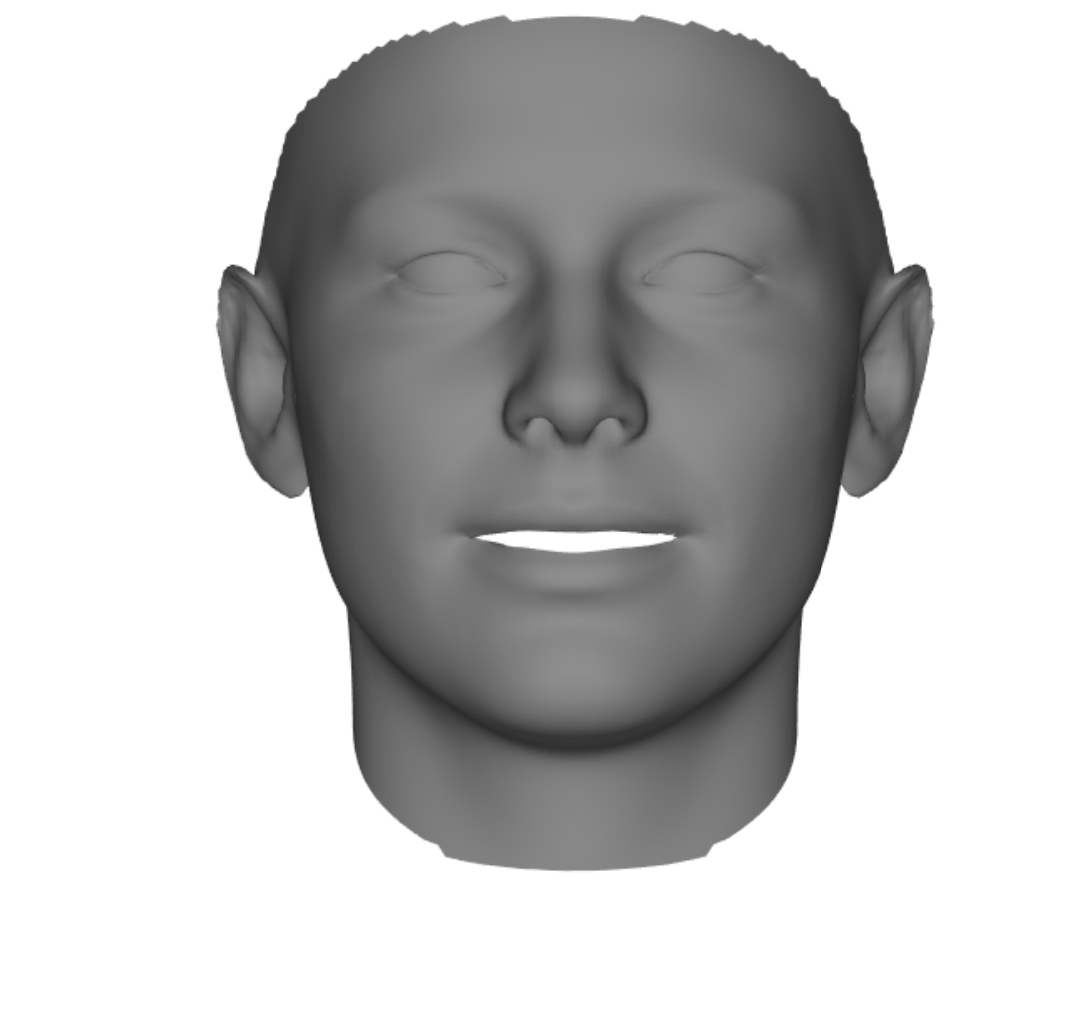} &
\includegraphics[width=\afifthcolumn]{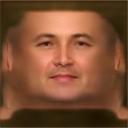} &
\includegraphics[trim=150 250 150 250,clip, width=\afifthcolumn]{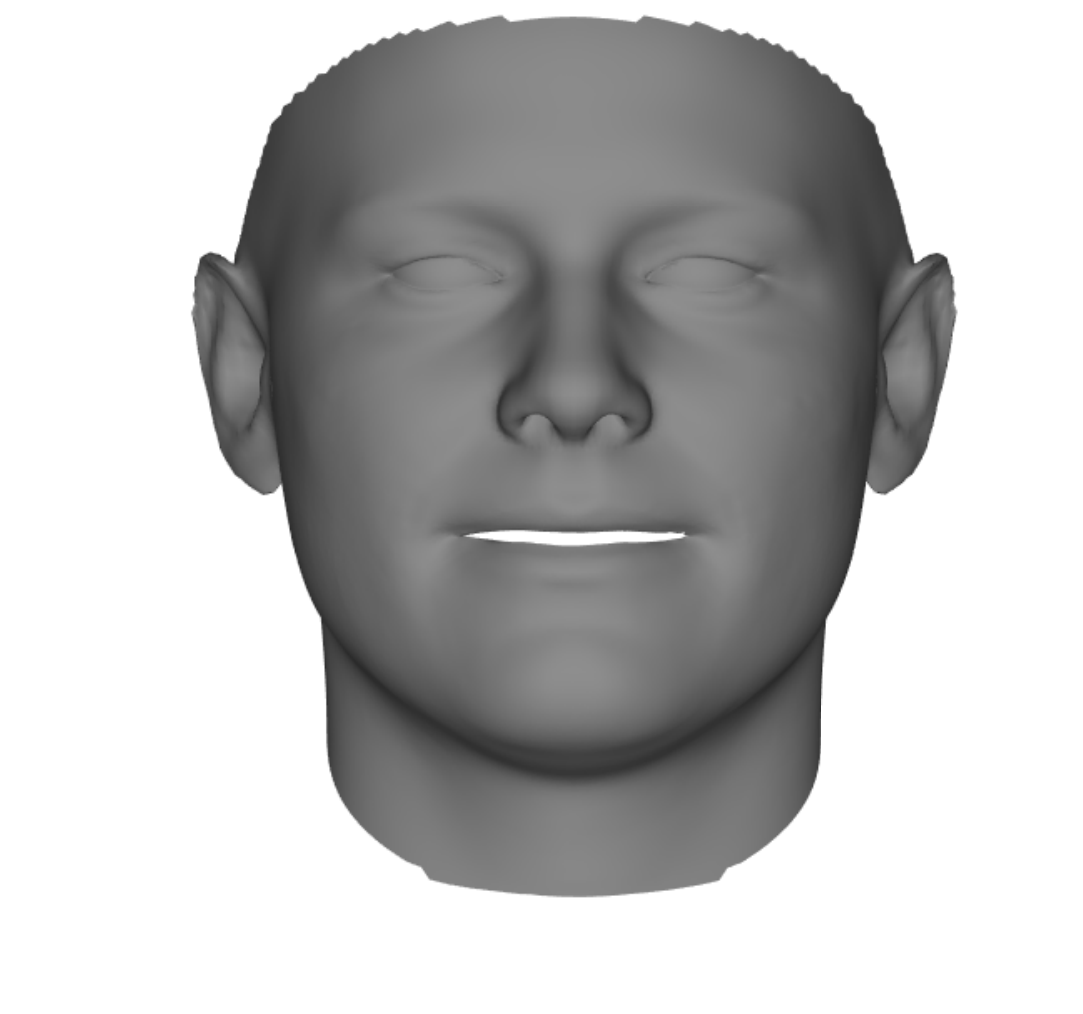} &
\includegraphics[width=\afifthcolumn]{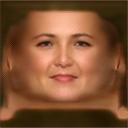} &
\includegraphics[trim=150 250 175 250,clip, width=\afifthcolumn]{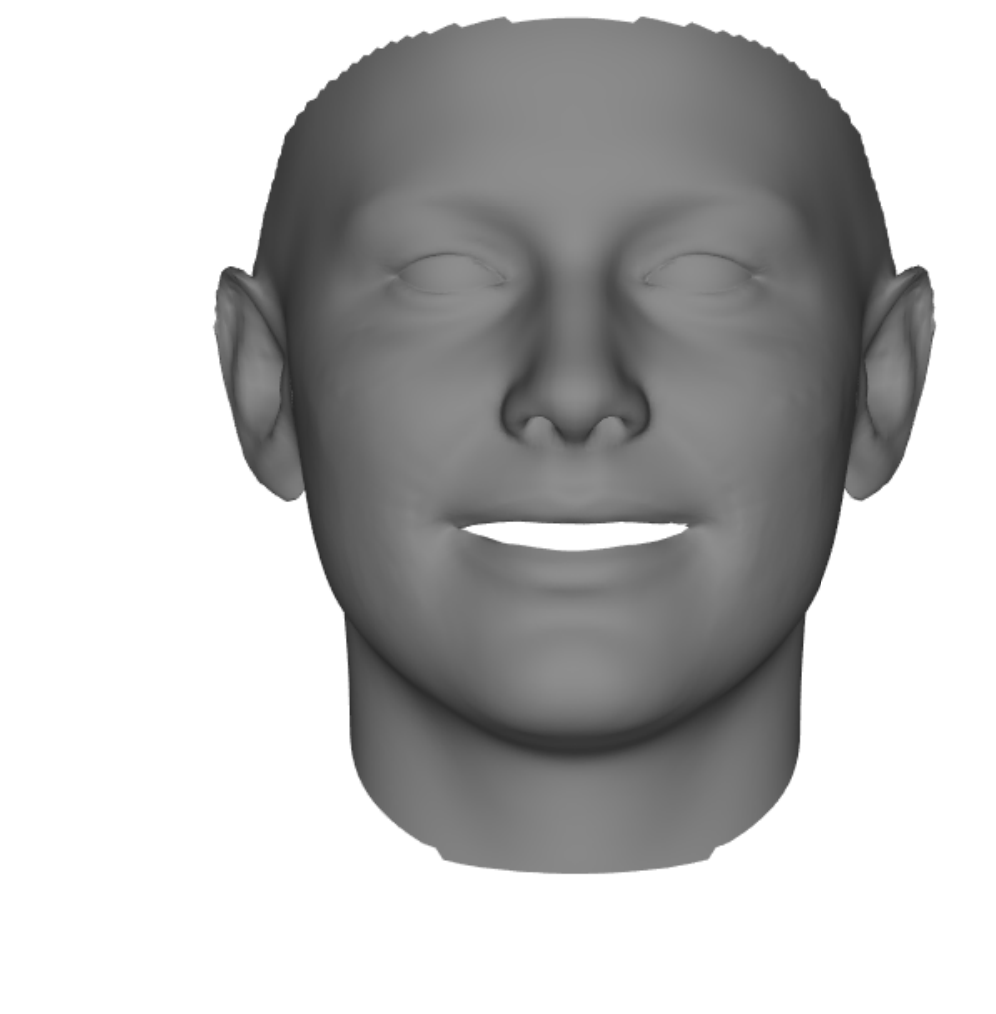} &
\\
\multicolumn{2}{c}{Female} & \multicolumn{2}{c}{Chubby} & \multicolumn{2}{c}{Smiling} \\
\end{tabular}
\vspace{-2mm}
\caption{\small Nonlinear 3DMM generates shape and texture embedded with different attributes.}
\label{fig:meaningful_basis}\figvspace \vspace{-2mm}
\end{center}
\end{figure}

\Paragraph{Attribute Embedding}
To better understand different shape and texture instances embedded in our two decoders, we dig into their attribute meaning.
For a given attribute, e.g., male, we feed images with that attribute $\{\mathbf{I}_i\}_{i=1}^n$ into our encoder to obtain two sets of parameters $\{\mathbf{f}_S^{i}\}_{i=1}^n$ and $\{\mathbf{f}_T^{i}\}_{i=1}^n$. 
These sets represent corresponding empirical distributions of the data in the low dimensional spaces. 
By computing the mean parameters $\mathbf{\bar{f}}_S, \mathbf{\bar{f}}_T$, and feed into their respective decoders, we can reconstruct the mean shape and texture with that attribute. Fig.~\ref{fig:meaningful_basis} visualizes the reconstructed shape and texture related to some attributes. 
Differences among attributes present in both shape and texture.

\begin{figure}[t!]
\begin{center}
\small
\setlength{\tabcolsep}{3pt}
\begin{tabular}{ @{}c@{}c@{}c@{}c@{}c@{}c@{}c@{\hskip 1.5mm}c@{}}
\multirow{2}{*}{Input} & \multirow{2}{*}{Linear} & \multicolumn{2}{c}{Nonlinear} \\ & & Grad desc & Network \\
\includegraphics[width=\afifthcolumn]{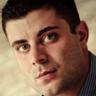} &
\includegraphics[trim=7 7 7 7,clip,width=\afifthcolumn]{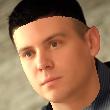} &
\includegraphics[width=\afifthcolumn]{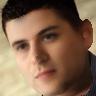} &
\includegraphics[width=\afifthcolumn]{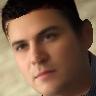} &
\\
\includegraphics[width=\afifthcolumn]{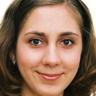} &
\includegraphics[trim=7 7 7 7,clip,width=\afifthcolumn]{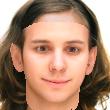} &
\includegraphics[width=\afifthcolumn]{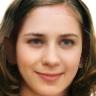} &
\includegraphics[width=\afifthcolumn]{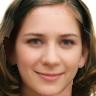} &
\\
\includegraphics[width=\afifthcolumn]{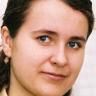} &
\includegraphics[trim=7 7 7 7,clip,width=\afifthcolumn]{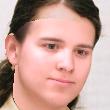} &
\includegraphics[width=\afifthcolumn]{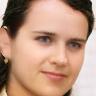} &
\includegraphics[width=\afifthcolumn]{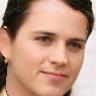} &
\\
\end{tabular}
\vspace{-2mm}
\caption{\small Texture representation power comparison. Our nonlinear model can better reconstruct the facial texture.}
\label{fig:tex_representationpower}\figvspace
\end{center}
\end{figure}

\begin{table}[t!]
\footnotesize
\caption{\small{Quantitative comparison of texture representation power.}} 
\label{tab:tex_representation_tab}
\vspace{-6mm}
\begin{center}
\begin{tabular}{ cccccccc}
\toprule 
Method & Linear & Nonlinear w.~Grad De. & Nonlinear w.~Network \\ \midrule
$L_1$    & $0.103$ & $0.066$   & $0.066$ \\ 
\bottomrule
\end{tabular}
\end{center}
\figvspace\vspace{-2mm}
\end{table}

\SubSection{Representation Power}
\Paragraph{Texture}
Given a face image, assuming we know the groundtruth shape and projection parameters, we can unwarp the texture into the UV space, as we generate ``pseudo groundtruth" texture in the weakly supervised step. 
With the groundtruth texture, by using gradient descent, we can estimate a texture parameter  $\mathbf{f}_T$ whose decoded texture matches with the groundtruth. 
Alternatively, we can minimize the reconstruction error in the image space, through the rendering layer with the groundtruth $\mathbf{S}$ and $\mathbf{m}$. 
Empirically, the two methods give similar performances but we choose the first option as it involves only one warping step, instead of rendering in every optimization iteration.
For the linear model, we use the fitting results of Basel texture and Phong illumination model~\cite{phong1975illumination} given by~\cite{zhu2016face}. 
As in Fig.~\ref{fig:tex_representationpower}, our nonlinear texture is closer to the groundtruth than the linear model, especially for in-the-wild images (the first two rows).
This is expected since the linear model is trained with controlled images. 
Quantitatively, our nonlinear model has significantly lower $L_1$ reconstruction error than the linear model ($0.066$ vs.~$0.103$, as in Tab.~\ref{tab:tex_representation_tab}).

\begin{figure}[t!]
\begin{center}
\small
\includegraphics[trim = 0 0 0 0, clip, width=0.95\linewidth]{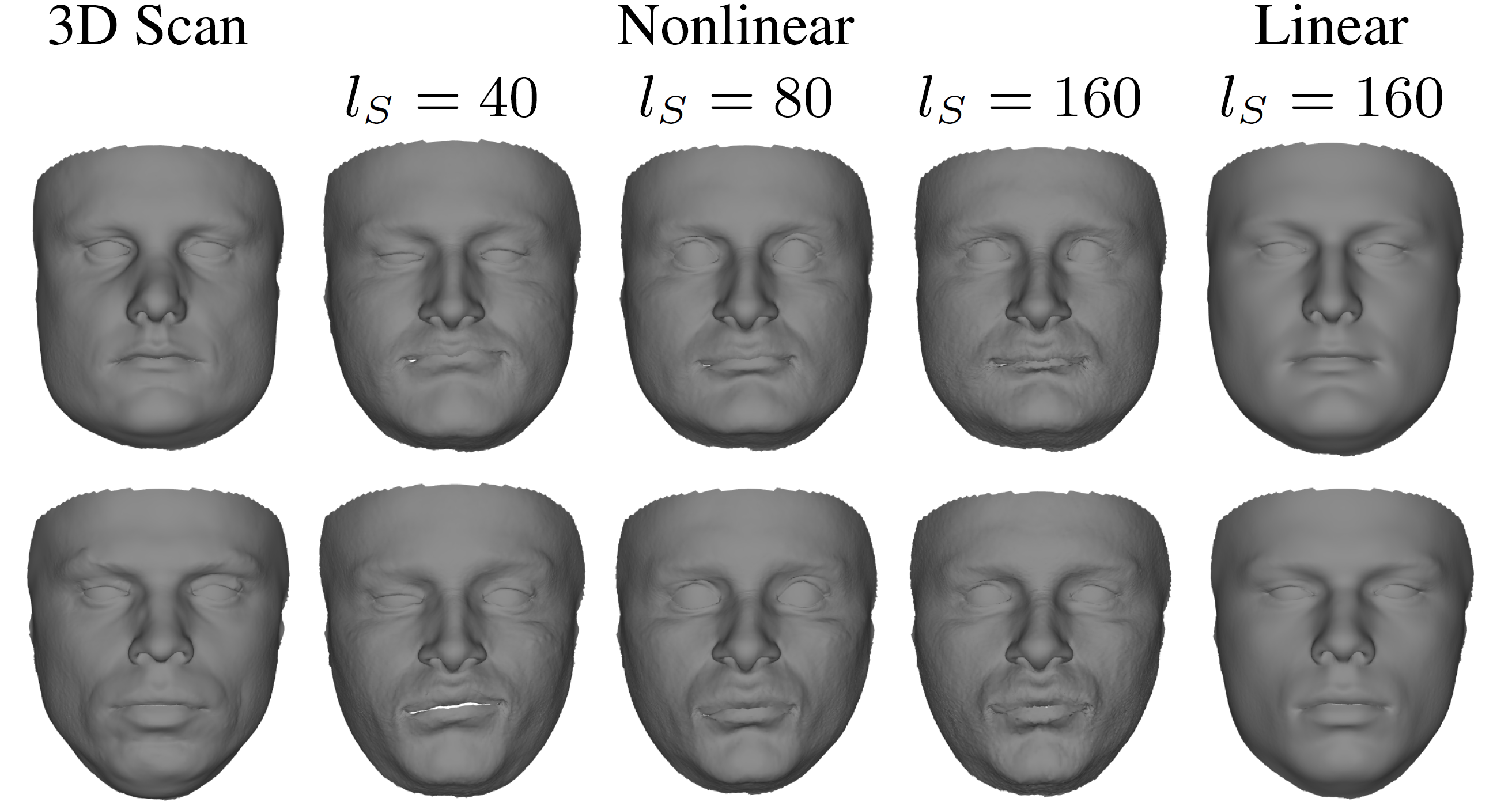}
\figvspace\vspace{1mm}
\caption{\small Shape representation power comparison.}
\label{fig:shape_representation}\figvspace
\end{center}
\end{figure}

\Paragraph{3D Shape}
We also compare the power of nonlinear and linear 3DMM in representing real-world 3D scans. 
We compare with BFM~\cite{paysan20093d}, the most commonly used 3DMM at present. 
We use ten 3D face scans provided by~\cite{paysan20093d}, which are not included in the training set of BFM.
As these face meshes are already registered using the same triangle definition with BFM,  no registration is necessary.
Given the groundtruth shape, by using gradient descent, we can estimate a shape parameter whose decoded shape matches the groundtruth. 
We define matching criteria on both vertex distances and surface normal direction. 
This empirically improves fidelity of final results compared to only optimizing vertex distances.
Also, to emphasize the compactness of nonlinear models, we train different models with different latent space sizes.
Fig.~\ref{fig:shape_representation} shows the visual quality of two models' reconstructions.
As we can see, our reconstructions closely match the face shapes. 
Meanwhile the linear model struggles with face shapes outside its PCA span.

To quantify the difference, we use NME, averaged per-vertex errors between the recovered and groundtruth shapes, normalized by inter-ocular distances. 
Our nonlinear model has a significantly smaller reconstruction error than the linear model, $0.0196$ vs.~$0.0241$ (Tab.~\ref{tab:shape_representation_number}). 
Also, the non-linear models are more compact. 
They can achieve similar performances as linear models whose latent space’s sizes doubled.

\begin{table}[t!]
\footnotesize
\caption{\small{3D scan reconstruction comparison (NME).}} 
\label{tab:shape_representation_number}
\vspace{-6mm}
\begin{center}
\begin{tabular}{ cccccccc}
\toprule 
$l_S$     & $40$ & $80$ & $160$ \\ \midrule
Linear    & $0.0321$ & $0.0279$ & $0.0241$ \\
Nonlinear & $0.0277$ & $0.0236$ & $\mathbf{0.0196}$ \\
\bottomrule
\end{tabular}
\end{center}
\figvspace
\end{table}

\begin{figure}[t!]
\begin{center}
\small
\setlength{\tabcolsep}{3pt}
\begin{tabular}{ @{\hskip 1.5mm}c@{\hskip 1.5mm}c@{\hskip 1.5mm}c@{\hskip 1.5mm}c@{}c@{}c@{}c@{\hskip 1.5mm}c@{}}
Input & Shape & Texture & Reconstruction \\
\includegraphics[width=\FittingFigWid]{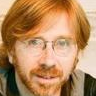} &
\includegraphics[trim=3 3 3 3,clip,width=\FittingFigShapeWid]{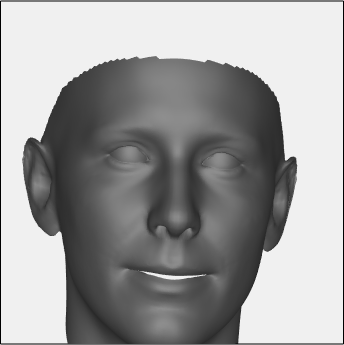} &
\includegraphics[width=\FittingFigWid]{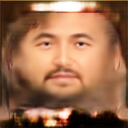} &
\includegraphics[width=\FittingFigWid]{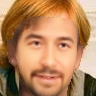} &
\\
\includegraphics[width=\FittingFigWid]{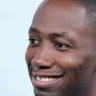} &
\includegraphics[trim=3 3 3 3,clip,width=\FittingFigShapeWid]{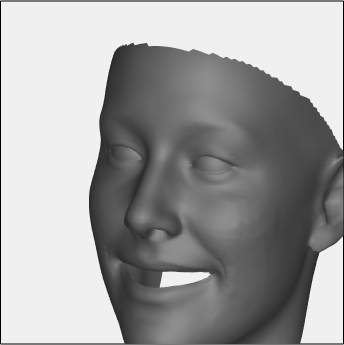} &
\includegraphics[width=\FittingFigWid]{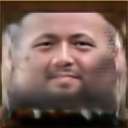} &
\includegraphics[width=\FittingFigWid]{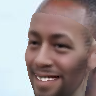} &
\\
\includegraphics[width=\FittingFigWid]{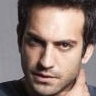} &
\includegraphics[trim=3 3 3 3,clip,width=\FittingFigShapeWid]{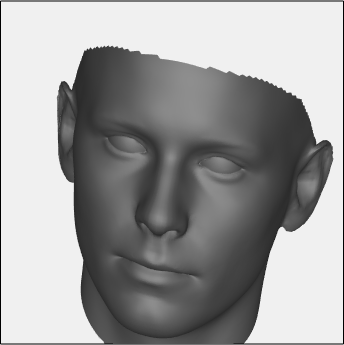} &
\includegraphics[width=\FittingFigWid]{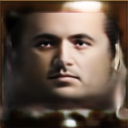} &
\includegraphics[width=\FittingFigWid]{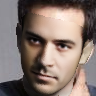} &
\\
\includegraphics[width=\FittingFigWid]{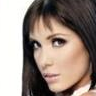} &
\includegraphics[trim=3 3 3 3,clip,width=\FittingFigShapeWid]{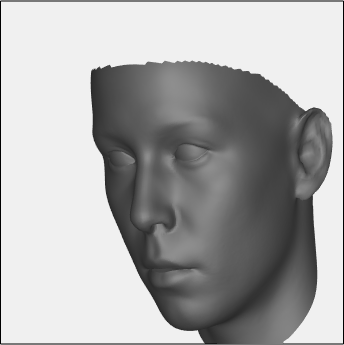} &
\includegraphics[width=\FittingFigWid]{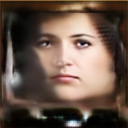} &
\includegraphics[width=\FittingFigWid]{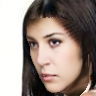} &
\\
\end{tabular}
\vspace{-2mm}
\caption{\small 3DMM fits to faces with diverse skin color, pose, expression, lighting, facial hair, and faithfully recovers these cues.}
\label{fig:3dmm_fitting}\figvspace
\end{center}
\end{figure}

\begin{figure}[t!]
\begin{center}
\small
\setlength{\tabcolsep}{3pt}
\begin{tabular}{ @{\hskip 1.5mm}c@{\hskip 1.5mm}c@{\hskip 1.5mm}c@{\hskip 1.5mm}c@{}c@{}c@{}c@{\hskip 1.5mm}c@{}}
\includegraphics[width=\AlignFigWid]{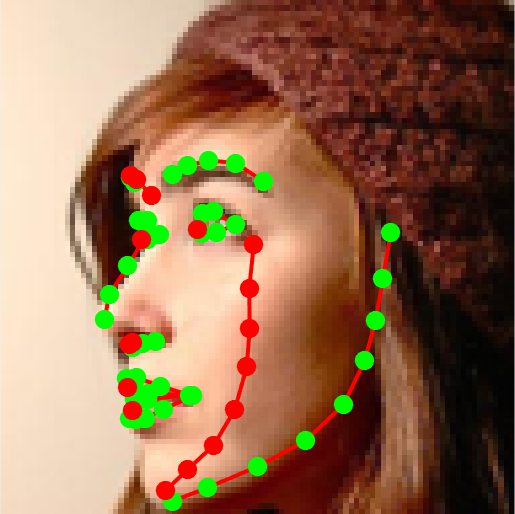} &
\includegraphics[width=\AlignFigWid]{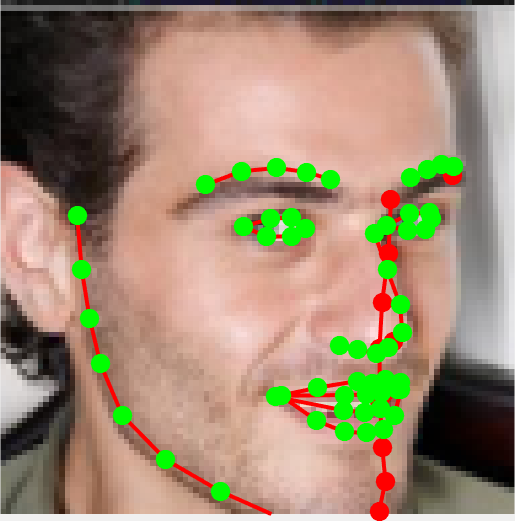} &
\includegraphics[width=\AlignFigWid]{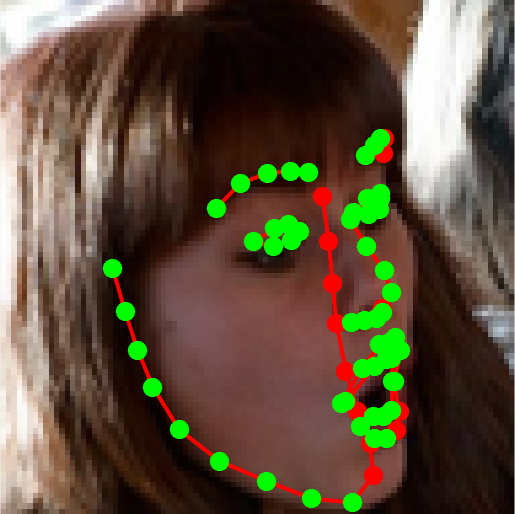} &
\includegraphics[width=\AlignFigWid]{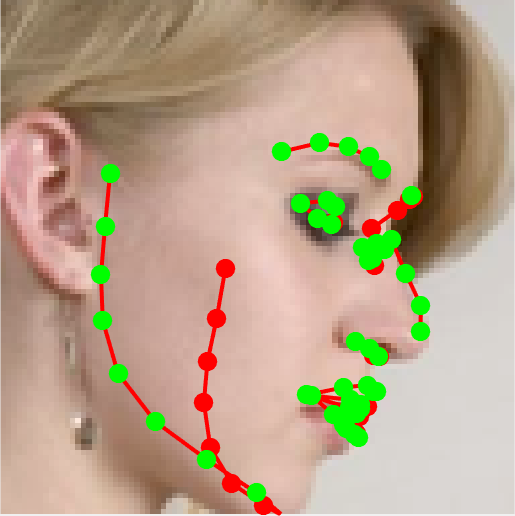} &
\\
\includegraphics[width=\AlignFigWid]{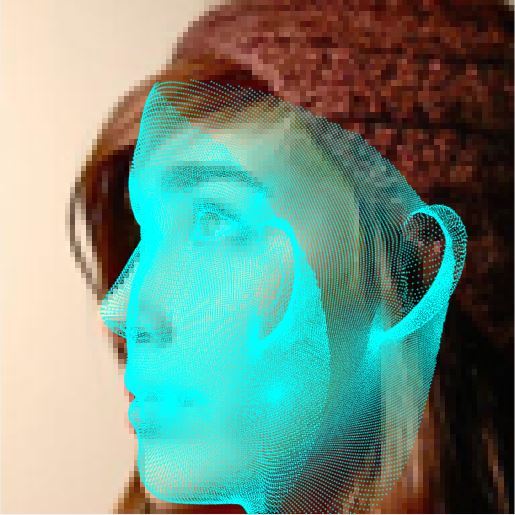} &
\includegraphics[width=\AlignFigWid]{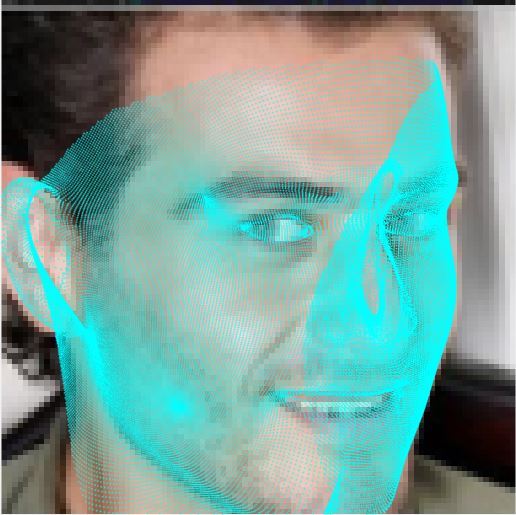} &
\includegraphics[width=\AlignFigWid]{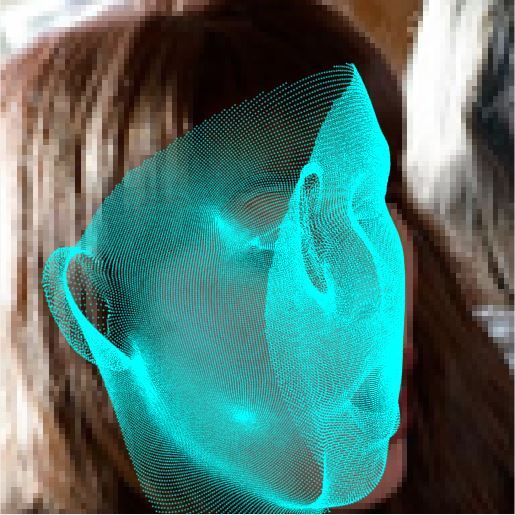} &
\includegraphics[width=\AlignFigWid]{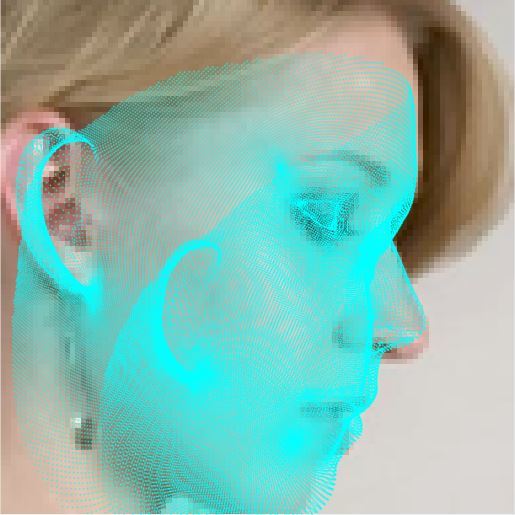} &
\\
\end{tabular}
\vspace{-2mm}
\caption{\small Our 2D face alignment results. Invisible landmarks are marked as red. We can handle extreme pose and/or expression. }
\label{fig:2d_align}\figvspace \vspace{-2mm}
\end{center}
\end{figure}

\begin{table}[t!]
\footnotesize
\caption{\small{Face alignment performance on ALFW2000}} 
\label{tab:2d_face_align}
\vspace{-6mm}
\begin{center}
\begin{tabular}{ cccccccc}
\toprule 
Method & Linear & SDM~\cite{yan2013learn} & 3DDFA~\cite{zhu2016face} & Ours \\ \midrule
NME    & $5.61$ & $6.12$                  & $5.42$                   & $\mathbf{4.70}$ \\ 
\bottomrule
\end{tabular}
\end{center}
\figvspace\vspace{-2mm}
\end{table}

\begin{figure*}[t!]
\begin{center}
\small
\setlength{\tabcolsep}{3pt}
\begin{tabular}{c@{\hskip 1.5mm}c@{\hskip 1mm}c@{\hskip 1.5mm}c@{\hskip 1mm}c@{\hskip 1.5mm}c@{\hskip 1mm}c@{\hskip 1mm}c@{}}
Input & \multicolumn{2}{c}{Our}  & \multicolumn{2}{c}{Richardson16} & \multicolumn{2}{c}{Tewari17} \\
\includegraphics[width=\FittingFigWid]{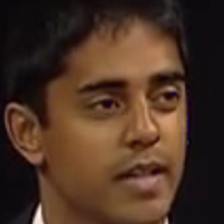} &
\includegraphics[width=\FittingFigWid]{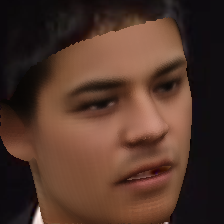} &
\includegraphics[width=\FittingFigWid]{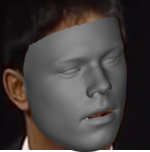} &
\includegraphics[width=\FittingFigWid]{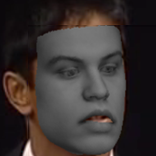} &
\includegraphics[width=\FittingFigWid]{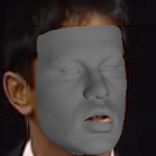} &
\includegraphics[width=\FittingFigWid]{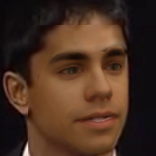} &
\includegraphics[width=\FittingFigWid]{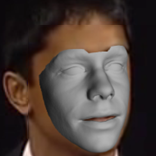} &
\\
\includegraphics[width=\FittingFigWid]{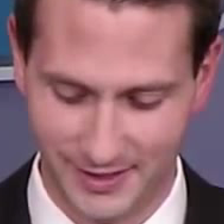} &
\includegraphics[width=\FittingFigWid]{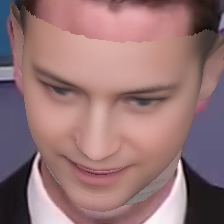} &
\includegraphics[width=\FittingFigWid]{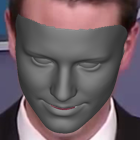} &
\includegraphics[width=\FittingFigWid]{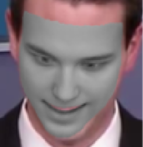} &
\includegraphics[width=\FittingFigWid]{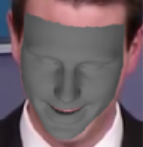} &
\includegraphics[width=\FittingFigWid]{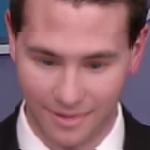} &
\includegraphics[width=\FittingFigWid]{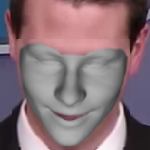} &
\\
\end{tabular}
\vspace{-2mm}
\caption{\small 3D reconstruction results comparison. We achieve comparable visual quality in 3D reconstruction.}
\label{fig:3drecon_exp}\figvspace \vspace{-2mm}
\end{center}
\end{figure*}

\begin{figure}[t!]
\centering
\includegraphics[trim=10 22 25 12,clip,width=\linewidth]{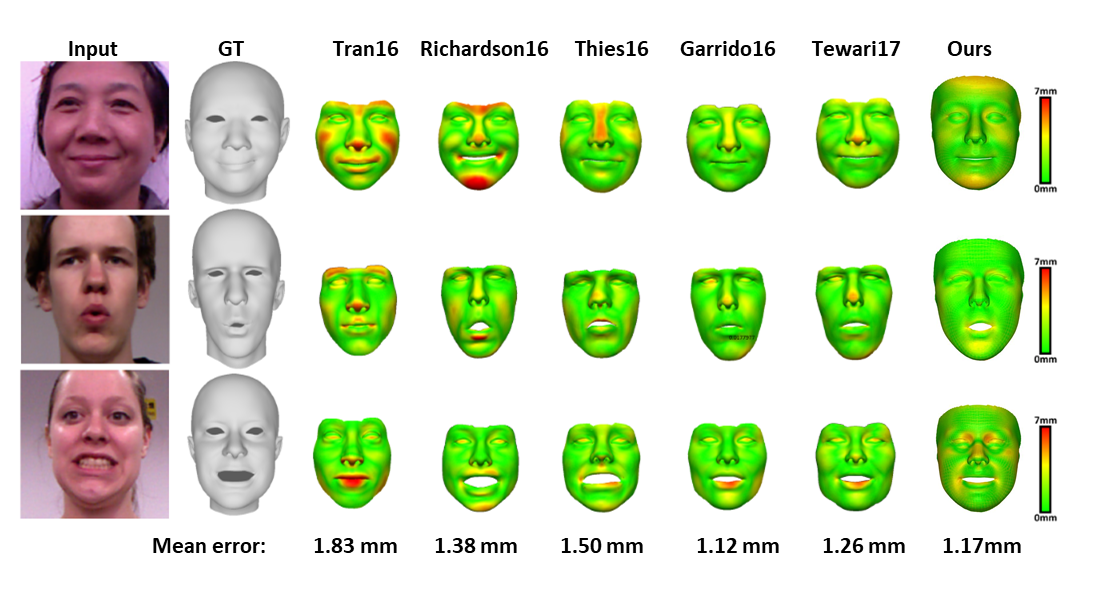}
\vspace{-4mm}
\caption{\small Quantitative evaluation of 3D reconstruction. We obtain a low error that is comparable to optimization-based methods.}
\label{fig:3d_rescon_quan}
\figvspace 
\end{figure}

\SubSection{Applications}
Having shown the capability of our nonlinear 3DMM (i.e., two decoders), now we demonstrate the applications of our entire network, which has the additional encoder.
Many applications of 3DMM are centered on its ability to fit to 2D face images.
Fig.~\ref{fig:3dmm_fitting} visualizes our 3DMM fitting results on CelebA dataset. 
Our encoder estimates the shape $\mathbf{S}$, texture $\mathbf{T}$ as well as projection parameter $\mathbf{m}$. 
We can recover personal facial characteristic in both shape and texture. 
Our texture can have variety skin color or facial hair, which is normally hard to be recovered by linear 3DMM.

\Paragraph{2D Face Alignment}
Face alignment is a critical step for any facial analysis task such as face recognition. 
With enhancement in the modeling, we hope to improve this task (Fig.~\ref{fig:2d_align}). 
We compare face alignment performance with state-of-the-art methods, SDM~\cite{yan2013learn} and 3DDFA~\cite{zhu2016face}, on the AFLW2000 dataset. 
The alignment accuracy is evaluated by the Normalized Mean Error (NME), the average of visible landmark error normalized by the bounding box size.
Here, current state-of-the-art 3DDFA~\cite{zhu2016face} is a cascade of CNNs that iteratively refines its estimation in multiple steps, meanwhile ours is a single-pass of $E$ and $D_S$. 
However, by jointly learning model fitting with 3DMM, our network can surpass ~\cite{zhu2016face}'s performance, as in Tab.~\ref{tab:2d_face_align}.
Another perspective is that in conventional 3DMM fitting~\cite{zhu2016face}, the texture is used as the input to regress the shape parameter, while ours adopts an analysis-by-synthesis scheme and texture is the output of the synthesis.
Further, for a more fair comparison of nonlinear vs.~linear models, we train an encoder with the same architecture as our $E$, whose output parameter will multiple with the linear shape bases $\mathbf{A}$, and train with the landmark loss function (Eqn.~\ref{eq:landmarkloss}). 
Again we observe the higher error from the linear model-based fitting.

\Paragraph{3D Face Reconstruction}
We compare our approach to recent works: the CNN-based iterative supervised regressor of Richardson et al.~\cite{richardson20163d, richardson2017learning} and unsupervised regressor method of Tewari et al.~\cite{tewari2017mofa}.
The work by Tewari et al.~\cite{tewari2017mofa} is relevant to us as they also learn to fit 3DMM in an unsupervised fashion. 
However, they are limited to linear 3DMM bases, which of course are not jointly trained with the model. 
Also, we only compare with the coarse network in~\cite{richardson2017learning} as their refinement network use SfS, which leads to a 2.5D representation and loses correspondence between different 3D shapes. 
This is orthogonal to our approach.
Fig.~\ref{fig:3drecon_exp} shows visual comparison. 
Following the same setting in~\cite{tewari2017mofa}, we also quantitatively compare our method with prior works on $9$ subjects of FaceWarehouse database~(Fig.~\ref{fig:3d_rescon_quan}). 
We achieve on-par results with Garrido et al.~\cite{garrido2016reconstruction}, an offline optimization method, while surpassing all other regression methods~\cite{tran2017regressing, richardson2017learning, tewari2017mofa}.

\begin{figure}[t!]
\begin{center}
\small
\setlength{\tabcolsep}{3pt}
\begin{tabular}{c@{\hskip 1.5mm}c@{\hskip 1mm}c@{\hskip 1.5mm}c@{\hskip 1mm}c@{\hskip 1.5mm}c@{\hskip 1mm}c@{\hskip 1mm}c@{}}
Input & No GAN & ImgGAN& PatchGAN \\
\includegraphics[width=\AblTexFigWid]{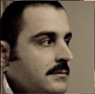} &
\includegraphics[width=\AblTexFigWid]{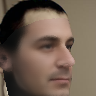} &
\includegraphics[width=\AblTexFigWid]{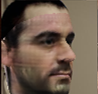} &
\includegraphics[width=\AblTexFigWid]{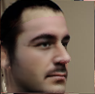} &
\\
\includegraphics[width=\AblTexFigWid]{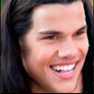} &
\includegraphics[width=\AblTexFigWid]{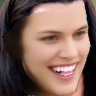} &
\includegraphics[width=\AblTexFigWid]{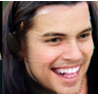} &
\includegraphics[width=\AblTexFigWid]{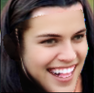} &
\end{tabular}
\vspace{-2mm}
\caption{\small Effects of adversarial losses for texture learning.}
\label{fig:abl_tex}
\figvspace \vspace{-2mm}
\end{center}
\end{figure}

\SubSection{Ablation on Texture Learning}
With great representation power, we would like to learn a realistic texture model from in-the-wild images. 
The rendering layer opens a possibility to apply adversarial loss in addition to global $L_1$ loss. 
Using a global image-based discriminator is redundant as the global structure is guaranteed by the rendering layer. 
Also, we empirically find that using global image-based discriminator can cause severe artifacts in the resultant texture. 
Fig.~\ref{fig:abl_tex} visualizes outputs of our network with different options of adversarial loss. 
Clearly, patchGAN offers higher realism and fewer artifacts.

\Section{Conclusions}
\label{sec:con}

Since its debut in 1999, 3DMM has became a cornerstone of facial analysis research 
with applications to many problems.
Despite its impact, it has drawbacks in requiring training data of 3D scans, learning from controlled 2D images, and limited representation power due to linear bases. 
These drawbacks could be formidable when fitting 3DMM to unconstrained faces, or learning 3DMM for generic objects such as shoes.
This paper demonstrates that there exists an alternative approach to 3DMM learning, where a nonlinear 3DMM can be learned from a large set of unconstrained face images without collecting 3D face scans.
Further, the model fitting algorithm can be learnt jointly with 3DMM, in an end-to-end fashion.

Our experiments cover a diverse aspects of our learnt model, some of which might need the subjective judgment of the readers.
We hope that both the judgment and quantitative results could be viewed under the context that, unlike linear 3DMM, no genuine 3D scans are used in our learning.
Finally, we believe that unsupervisedly learning 3D models from large-scale in-the-wild 2D images is one promising research direction. 
This work is one step along this direction.

{\small
\bibliographystyle{ieee}
\bibliography{egbib}

\begin{thebibliography}{10}\itemsep=-1pt

\bibitem{aldrian2013inverse}
O.~Aldrian and W.~A. Smith.
\newblock Inverse rendering of faces with a 3{D} morphable model.
\newblock {\em TPAMI}, 2013.

\bibitem{amberg2008expression}
B.~Amberg, R.~Knothe, and T.~Vetter.
\newblock Expression invariant 3{D} face recognition with a morphable model.
\newblock In {\em FG}, 2008.

\bibitem{amberg2007optimal}
B.~Amberg, S.~Romdhani, and T.~Vetter.
\newblock Optimal step nonrigid {ICP} algorithms for surface registration.
\newblock In {\em CVPR}, 2007.

\bibitem{blanz1999morphable}
V.~Blanz and T.~Vetter.
\newblock A morphable model for the synthesis of 3{D} faces.
\newblock In {\em Proceedings of the 26th annual conference on Computer
  graphics and interactive techniques}. ACM Press/Addison-Wesley Publishing
  Co., 1999.

\bibitem{blanz2003face}
V.~Blanz and T.~Vetter.
\newblock Face recognition based on fitting a 3{D} morphable model.
\newblock {\em TPAMI}, 2003.

\bibitem{bolkart2015groupwise}
T.~Bolkart and S.~Wuhrer.
\newblock A groupwise multilinear correspondence optimization for 3{D} faces.
\newblock In {\em ICCV}, 2015.

\bibitem{booth20173d}
J.~Booth, E.~Antonakos, S.~Ploumpis, G.~Trigeorgis, Y.~Panagakis, and
  S.~Zafeiriou.
\newblock 3{D} face morphable models ``{I}n-the-wild''.
\newblock In {\em CVPR}, 2017.

\bibitem{booth20163d}
J.~Booth, A.~Roussos, S.~Zafeiriou, A.~Ponniah, and D.~Dunaway.
\newblock A 3{D} morphable model learnt from 10,000 faces.
\newblock In {\em CVPR}, 2016.

\bibitem{cao2014facewarehouse}
C.~Cao, Y.~Weng, S.~Zhou, Y.~Tong, and K.~Zhou.
\newblock Facewarehouse: A 3{D} facial expression database for visual
  computing.
\newblock {\em IEEE Transactions on Visualization and Computer Graphics}, 2014.

\bibitem{cole2017face}
F.~Cole, D.~Belanger, D.~Krishnan, A.~Sarna, I.~Mosseri, and W.~T. Freeman.
\newblock Face synthesis from facial identity features.
\newblock In {\em CVPR}, 2017.

\bibitem{cootes2001active}
T.~F. Cootes, G.~J. Edwards, and C.~J. Taylor.
\newblock Active appearance models.
\newblock {\em TPAMI}, 2001.

\bibitem{dollar2010cascaded}
P.~Doll{\'a}r, P.~Welinder, and P.~Perona.
\newblock Cascaded pose regression.
\newblock In {\em CVPR}, 2010.

\bibitem{garrido2016reconstruction}
P.~Garrido, M.~Zollh{\"o}fer, D.~Casas, L.~Valgaerts, K.~Varanasi,
  P.~P{\'e}rez, and C.~Theobalt.
\newblock Reconstruction of personalized 3{D} face rigs from monocular video.
\newblock {\em ACM Transactions on Graphics (TOG)}, 2016.

\bibitem{goodfellow2014generative}
I.~Goodfellow, J.~Pouget-Abadie, M.~Mirza, B.~Xu, D.~Warde-Farley, S.~Ozair,
  A.~Courville, and Y.~Bengio.
\newblock Generative adversarial nets.
\newblock In {\em NIPS}, 2014.

\bibitem{gu2008generative}
L.~Gu and T.~Kanade.
\newblock A generative shape regularization model for robust face alignment.
\newblock In {\em ECCV}, 2008.

\bibitem{jourabloo2015pose}
A.~Jourabloo and X.~Liu.
\newblock Pose-invariant 3{D} face alignment.
\newblock In {\em ICCV}, 2015.

\bibitem{jourabloo2016large}
A.~Jourabloo and X.~Liu.
\newblock Large-pose face alignment via {CNN}-based dense 3{D} model fitting.
\newblock In {\em CVPR}, 2016.

\bibitem{jourabloo2017poseinvariant}
A.~Jourabloo and X.~Liu.
\newblock Pose-invariant face alignment via {CNN}-based dense 3{D} model
  fitting.
\newblock {\em IJCV}, 2017.

\bibitem{jourabloo2017pose}
A.~Jourabloo, X.~Liu, M.~Ye, and L.~Ren.
\newblock Pose-invariant face alignment with a single {CNN}.
\newblock In {\em ICCV}, 2017.

\bibitem{koppen2017gaussian}
P.~Koppen, Z.-H. Feng, J.~Kittler, M.~Awais, W.~Christmas, X.-J. Wu, and H.-F.
  Yin.
\newblock Gaussian mixture 3{D} morphable face model.
\newblock {\em Pattern Recognition}, 2017.

\bibitem{liu2016joint}
F.~Liu, D.~Zeng, Q.~Zhao, and X.~Liu.
\newblock Joint face alignment and 3{D} face reconstruction.
\newblock In {\em ECCV}. Springer, 2016.

\bibitem{face-model-fitting-on-low-resolution-images}
X.~Liu, P.~Tu, and F.~Wheeler.
\newblock Face model fitting on low resolution images.
\newblock In {\em BMVC}, 2006.

\bibitem{liu2015faceattributes}
Z.~Liu, P.~Luo, X.~Wang, and X.~Tang.
\newblock Deep learning face attributes in the wild.
\newblock In {\em ICCV}, 2015.

\bibitem{mcdonagh2016joint}
J.~McDonagh and G.~Tzimiropoulos.
\newblock Joint face detection and alignment with a deformable {H}ough
  transform model.
\newblock In {\em ECCV}, 2016.

\bibitem{nhan2015beyond}
C.~Nhan~Duong, K.~Luu, K.~Gia~Quach, and T.~D. Bui.
\newblock Beyond principal components: Deep {B}oltzmann {M}achines for face
  modeling.
\newblock In {\em CVPR}, 2015.

\bibitem{paysan20093d}
P.~Paysan, R.~Knothe, B.~Amberg, S.~Romdhani, and T.~Vetter.
\newblock A 3{D} face model for pose and illumination invariant face
  recognition.
\newblock In {\em Advanced video and signal based surveillance, 2009. AVSS'09.
  Sixth IEEE International Conference on}. IEEE, 2009.

\bibitem{phong1975illumination}
B.~T. Phong.
\newblock Illumination for computer generated pictures.
\newblock {\em Communications of the ACM}, 1975.

\bibitem{radford2015unsupervised}
A.~Radford, L.~Metz, and S.~Chintala.
\newblock Unsupervised representation learning with deep convolutional
  generative adversarial networks.
\newblock In {\em ICLR}, 2016.

\bibitem{richardson20163d}
E.~Richardson, M.~Sela, and R.~Kimmel.
\newblock 3{D} face reconstruction by learning from synthetic data.
\newblock In {\em 3DV}. IEEE, 2016.

\bibitem{richardson2017learning}
E.~Richardson, M.~Sela, R.~Or-El, and R.~Kimmel.
\newblock Learning detailed face reconstruction from a single image.
\newblock In {\em CVPR}, 2017.

\bibitem{adaptive-3d-face-reconstruction-from-unconstrained-photo-collections}
J.~Roth, Y.~Tong, and X.~Liu.
\newblock Adaptive 3{D} face reconstruction from unconstrained photo
  collections.
\newblock {\em TPAMI}, 2017.

\bibitem{sagonas2016300}
C.~Sagonas, E.~Antonakos, G.~Tzimiropoulos, S.~Zafeiriou, and M.~Pantic.
\newblock 300 faces in-the-wild challenge: Database and results.
\newblock {\em Image and Vision Computing}, 2016.

\bibitem{shrivastava2017learning}
A.~Shrivastava, T.~Pfister, O.~Tuzel, J.~Susskind, W.~Wang, and R.~Webb.
\newblock Learning from simulated and unsupervised images through adversarial
  training.
\newblock In {\em CVPR}, 2017.

\bibitem{staal2015describing}
F.~C. Staal, A.~J. Ponniah, F.~Angullia, C.~Ruff, M.~J. Koudstaal, and
  D.~Dunaway.
\newblock Describing crouzon and pfeiffer syndrome based on principal component
  analysis.
\newblock {\em Journal of Cranio-Maxillofacial Surgery}, 2015.

\bibitem{tewari2017mofa}
A.~Tewari, M.~Zollh{\"o}fer, H.~Kim, P.~Garrido, F.~Bernard, P.~P{\'e}rez, and
  C.~Theobalt.
\newblock {MoFA}: Model-based deep convolutional face autoencoder for
  unsupervised monocular reconstruction.
\newblock In {\em ICCV}, 2017.

\bibitem{tran2017regressing}
A.~T. Tran, T.~Hassner, I.~Masi, and G.~Medioni.
\newblock Regressing robust and discriminative 3{D} morphable models with a
  very deep neural network.
\newblock In {\em CVPR}, 2017.

\bibitem{tran2017disentangled}
L.~Tran, X.~Yin, and X.~Liu.
\newblock Disentangled representation learning {GAN} for pose-invariant face
  recognition.
\newblock In {\em CVPR}, 2017.

\bibitem{tran2018representation}
L.~Tran, X.~Yin, and X.~Liu.
\newblock Representation learning by rotating your faces.
\newblock {\em TPAMI}, 2018.

\bibitem{tulyakov2015regressing}
S.~Tulyakov and N.~Sebe.
\newblock Regressing a 3{D} face shape from a single image.
\newblock In {\em ICCV}, 2015.

\bibitem{vlasic2005face}
D.~Vlasic, M.~Brand, H.~Pfister, and J.~Popovi{\'c}.
\newblock Face transfer with multilinear models.
\newblock In {\em ACM transactions on graphics (TOG)}. ACM, 2005.

\bibitem{wu2015robust}
Y.~Wu and Q.~Ji.
\newblock Robust facial landmark detection under significant head poses and
  occlusion.
\newblock In {\em ICCV}, 2015.

\bibitem{yan2013learn}
J.~Yan, Z.~Lei, D.~Yi, and S.~Li.
\newblock Learn to combine multiple hypotheses for accurate face alignment.
\newblock In {\em CVPRW}, 2013.

\bibitem{yin20063d}
L.~Yin, X.~Wei, Y.~Sun, J.~Wang, and M.~J. Rosato.
\newblock A 3{D} facial expression database for facial behavior research.
\newblock In {\em FGR}, 2006.

\bibitem{yin2017towards}
X.~Yin, X.~Yu, K.~Sohn, X.~Liu, and M.~Chandraker.
\newblock Towards large-pose face frontalization in the wild.
\newblock In {\em ICCV}, 2017.

\bibitem{zhang2006face}
L.~Zhang and D.~Samaras.
\newblock Face recognition from a single training image under arbitrary unknown
  lighting using spherical harmonics.
\newblock {\em TPAMI}, 2006.

\bibitem{zhu2016face}
X.~Zhu, Z.~Lei, X.~Liu, H.~Shi, and S.~Z. Li.
\newblock Face alignment across large poses: A 3{D} solution.
\newblock In {\em CVPR}, 2016.

\end{thebibliography}
}

\end{document}